\documentclass[10pt,twocolumn,letterpaper]{article}

\usepackage[pagenumbers]{iccv} 
\usepackage{multirow}

\definecolor{iccvblue}{rgb}{0.21,0.49,0.74}
\usepackage[pagebackref,breaklinks,colorlinks,allcolors=iccvblue]{hyperref}

\title{DLF: Extreme Image Compression with Dual-generative Latent Fusion}

\author{
Naifu Xue$^{1*}$ \quad 
Zhaoyang Jia$^2$\thanks{Naifu Xue and Zhaoyang Jia are  visiting students of MSRA.} \quad
Jiahao Li $^3$ \quad
Bin Li$^3$ \quad
Yuan Zhang $^1$ \quad
Yan Lu$^3$  \\
$^1$ Communication University of China \quad 
$^2$ University of Science and Technology of China\\
$^3$ Microsoft Research Asia\\
{\tt\small \{aaronxuenf, yzhang\}@cuc.edu.cn, \{jzy$\_$ustc\}@mail.ustc.edu.cn }\\
{\tt\small \{li.jiahao, libin, yanlu\}@microsoft.com}
}
\vspace{-3mm}

\begin{document}
\maketitle

\begin{abstract}
Recent studies in extreme image compression have achieved remarkable performance by compressing the tokens from generative tokenizers. 
However, these methods often prioritize clustering common semantics within the dataset, while overlooking the diverse details of individual objects.
Consequently, this results in suboptimal reconstruction fidelity, especially at low bitrates.
To address this issue, we introduce a Dual-generative Latent Fusion (DLF) paradigm. 
DLF decomposes the latent into semantic and detail elements, compressing them through two distinct branches.
The semantic branch clusters high-level information into compact tokens, while the detail branch encodes perceptually critical details to enhance overall fidelity.
Additionally, we propose a cross-branch interactive design to reduce redundancy between the two branches, thereby minimizing the overall bit cost. 
Experimental results demonstrate the impressive reconstruction quality of DLF even below 0.01 bits per pixel (bpp). 
On the CLIC2020 test set, our method achieves bitrate savings of up to 27.93\% on LPIPS and 53.55\% on DISTS compared to MS-ILLM. 
Furthermore, DLF surpasses recent diffusion-based codecs in visual fidelity while maintaining a comparable level of generative realism.
Project: \url{https://dlfcodec.github.io/}
\end{abstract}
\vspace{-3mm}

\begingroup
\renewcommand\thefootnote{}
\footnotetext{This paper stems from an open-source project initiated in March 2024.}
\endgroup

\section{Introduction}
\label{sec:intro}

\begin{figure}[t]
    \begin{center}
        \includegraphics[width=1.0\linewidth]{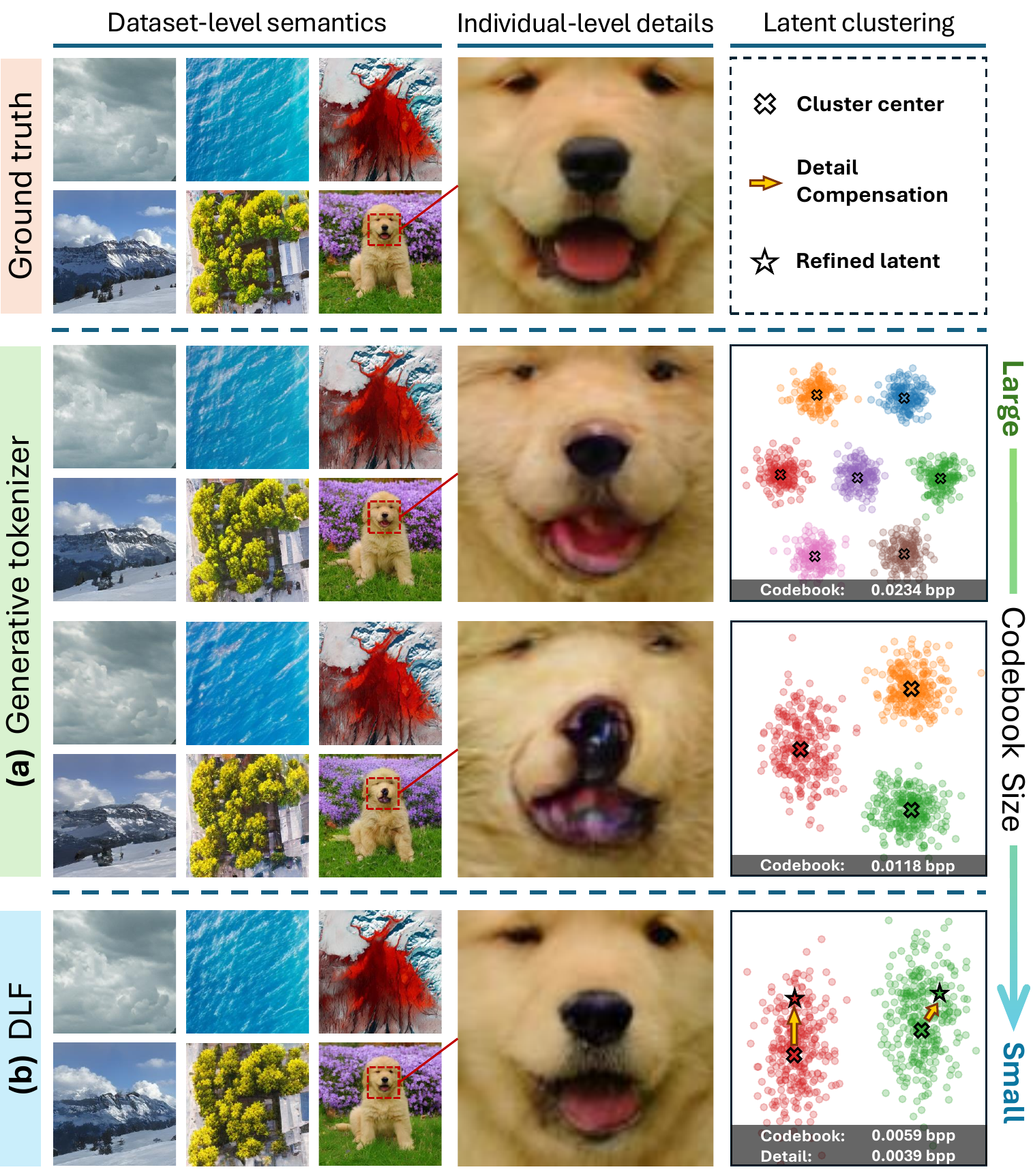}
    \end{center}
    \vspace{-6mm}
    \caption{
        Visualization of reconstruction and clustered latents. 
        (a) Generative tokenizer \cite{esser2021taming} clusters dataset-level semantics to encode common contents. But with reduced codebook size, the details in individual objects become significantly distorted.
        (b) Our DLF encodes the additional detail information to compensate for the small codebook, significantly improving overall fidelity yet with lower total bitrate.
    }
    \label{fig:intro_1}
    \vspace{-4mm}
\end{figure}

\begin{figure*}[t]
    \begin{center}
        \includegraphics[width=1.0\linewidth]{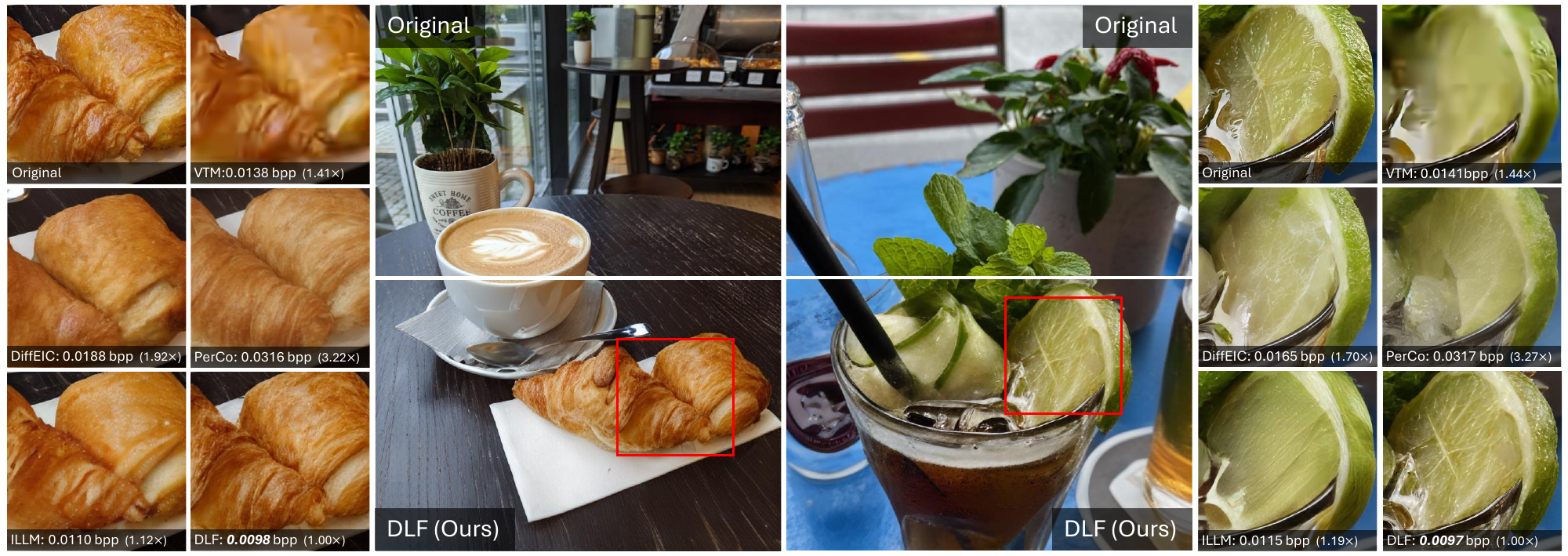}
    \end{center}
    \vspace{-6mm}
    \caption{
        Qualitative comparisons with VTM \cite{bross2021overview}, MS-ILLM \cite{muckley2023improving}, DiffEIC \cite{li2024extremeimagecompressionlatent} and PerCo \cite{careil2023towards}.
        Our DLF demonstrates superior fidelity at the lowest bitrate, even in regions with complex details. 
        Conversely, the VTM and MS-ILLM tend to produce over-smoothed images, while DiffEIC and PerCo struggle to maintain satisfactory fidelity.
    }
    \label{fig:intro_2}
    \vspace{-5mm}
\end{figure*}

\begin{figure}[t]
    \begin{center}
        \includegraphics[width=1.0\linewidth]{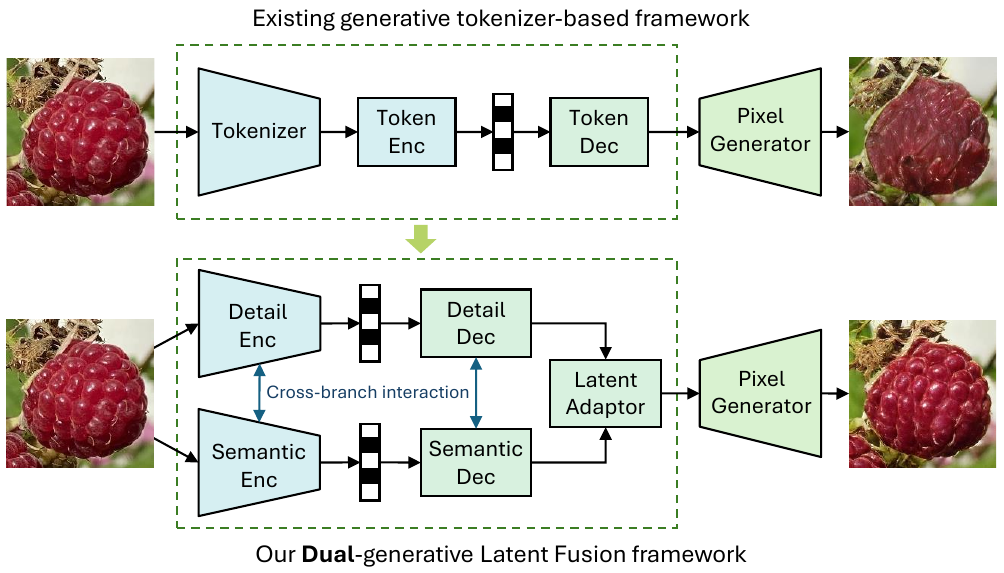}
    \end{center}
    \vspace{-5mm}
    \caption{
        Frameworks in \cite{mao2024extreme, xue2024unifying, Jia_2024_CVPR} and DLF.
        %
        Instead of directly compressing the tokenizer’s latent, our DLF splits it into the semantic and detail parts for flexible compression.
        %
        %
    }
    \label{fig:intro_3}
    \vspace{-5mm}
\end{figure}

As image data continues to grow, achieving efficient compression while maintaining human perceptual quality has become crucial. At low bitrates, both traditional codecs \cite{bross2021overview, choi2020overview} and learning-based codecs \cite{Cheng_2020_CVPR, jiang2023mlic, Liu_2023_CVPR} (optimized for PSNR) often produce blurry reconstructions due to substantial information degradation \cite{blau2019rethinking}. In this context, generative codecs \cite{Agustsson_2019_ICCV, mentzer2020high, muckley2023improving}, which enhance subjective quality by adding high-frequency details to reconstructed images, have attracted considerable attention. To achieve effective generative coding at extremely low bitrates, recent studies \cite{mao2024extreme, xue2024unifying, Jia_2024_CVPR, qi2025generative} have employed generative visual tokenizers \cite{van2017neural, esser2021taming} as foundational components. These tokenizers cluster common semantic content across the entire dataset into a compact codebook while simultaneously learning a powerful pixel generator based on the clustered information. As illustrated in Fig.~\ref{fig:intro_1} (a), if you do not look closely, many common semantic content within the dataset can be reconstructed with acceptable quality using a small codebook. Consequently, such codecs \cite{mao2024extreme, xue2024unifying, Jia_2024_CVPR} only need to compress and transmit a few token indices, thereby significantly enhancing the compression ratio.
However, when we zoom in each image to focus on the details of individual objects, we observe that the limited codebook capacity fails to capture these details accurately when compared with the ground truth. This inadequacy is particularly unfavorable for objects with precise geometric features, leading to a noticeable degradation in fidelity even without a reference for comparison. As illustrated in Fig.~\ref{fig:intro_1} (a), a reduced number of clusters results in severe distortion on a dog's face. This observation underscores the limitation of the aforementioned codecs \cite{mao2024extreme, xue2024unifying, Jia_2024_CVPR}: their backend tokenizer employs a single codebook to learn uniform latent clusters for all image content, failing to balance the demand between the common semantic learning across the dataset and the diverse detail representation of the individual objects. Therefore, the question arises: \textit{How can we compress the latent in a more flexible manner, dynamically adapting to the content variations of each image?}
%


%
To tackle this challenge, we introduce a novel extreme image coding paradigm named Dual-generative Latent Fusion (DLF). As illustrated in Fig.~\ref{fig:intro_3}, DLF decomposes the image latent into semantic and detail components, employing two parallel branches to compress each component according to its demands. The semantic branch inherits the clustering capability of the generative tokenizer, capturing the common content at the dataset-level with few bits. Meanwhile, the detail branch complements the shortcomings of the single codebook tokenizer, representing the diverse details through a large quantization space. By training with a rate-distortion loss, the detail branch learns to adaptively allocate more bits to the most critical regions for enhanced fidelity, while avoiding the excessive data overhead. Finally, the decoded semantic and detail features are fused using a latent adaptor, aligning them into the same latent space for high-quality reconstruction using the pixel generator. As shown in Fig.~\ref{fig:intro_1} (b), our DLF not only reduces bit cost but also significantly enhances the detail fidelity while preserving the overall perceptual quality.
%

%


%
By adopting a dual-branch coding design, DLF generates two bitstreams, similar to HybridFlow \cite{lu2024hybridflow}. To minimize the overall bit cost, it is crucial to reduce redundancy between the two bitstreams. To this end, we introduce a cross-branch interaction mechanism that employs multiple attention-based layers to jointly process the intermediate features of both branches. This interaction not only optimizes bit allocation between the bottlenecks but also enables each branch to retain only the complementary information, thereby considerably reducing redundancy. By contrast, HybridFlow adopts two independent encoders without joint design, increasing bit cost due to redundancy.

Powered by our advanced designs, DLF achieves state-of-the-art (SOTA) compression performance across various datasets and metrics, especially at extremely low bitrates such as 0.01 bpp. Qualitative results in Fig.~\ref{fig:intro_2} highlight the superiority of our method in terms of both generative realism and reconstruction fidelity. Compared to MS-ILLM \cite{muckley2023improving}, our approach achieves bitrate savings of 27.93\% for LPIPS and 53.55\% for DISTS on the CLIC2020 dataset. Additionally, our DLF significantly surpasses recent diffusion-based codecs \cite{careil2023towards, li2024extremeimagecompressionlatent, li2024diffusionbasedextremeimagecompression} in terms of visual fidelity while maintaining comparable realism.
In summary, our contributions are as follows:
\begin{itemize}
    \item We propose a novel dual-branch coding framework for extreme image compression. It decomposes features into general semantic and specific detail parts, dynamically adapting to the content variations of each individual image and enabling flexible compression.
    \item We propose a cross-branch interactive structure to eliminate redundancy between the semantic and detail bitstreams, further facilitating compression efficiency.
    \item DLF achieves SOTA performance on various datasets and metrics, delivering both high realism and faithful reconstruction at extremely low bitrates. 
\end{itemize}

\section{Related work}
\subsection{Learned Image Compression} 
Variational autoencoder (VAE)-based image codecs \cite{jiang2023mlic, Liu_2023_CVPR, wang2023evc, jin2024neural, zhao2021universal, zhao2023universal} have been widely studied. However, these methods, mainly optimized for PSNR, produce blurry images at low bitrates due to severe detail loss \cite{blau2019rethinking}.
To enhance perceptual quality, Agustsson et al. \cite{Agustsson_2019_ICCV} introduced the concept of generative image compression, using GANs \cite{goodfellow2014generative} to generate the lost details.
Mentzer et al. \cite{mentzer2020high} proposed HiFiC, which features an advanced network and a generative adversarial loss.
The recent MS-ILLM \cite{muckley2023improving} adopts a non-binary discriminator to greatly enhance statistical fidelity.
Lee et al.\cite{lee2024neuralimagecompressiontextguided} further utilize text features to assist the encoder, thus improving semantic consistency.
Despite these advancements, existing VAE-based generative image codecs \cite{Agustsson_2019_ICCV, mentzer2020high, muckley2023improving, lee2024neuralimagecompressiontextguided} struggle to deliver pleasing quality at extremely low bitrates. 
Consequently, recent work has begun exploring the use of generative visual tokenizers \cite{esser2021taming} and diffusion models \cite{rombach2021highresolution} for extreme image compression.
\subsection{Tokenizer-based Extreme Image Compression} 
Image tokenizers with discrete codebooks, such as VQVAE \cite{van2017neural} and its subsequent work \cite{razavi2019generating}, can learn semantically rich tokens through latent clustering. The extracted tokens can be utilized for reconstruction, generation \cite{esser2021taming, chang2022maskgit}, and other visual tasks \cite{rao2021dynamicvit, Yin_2022_CVPR, yu2024spae}. The representative generative tokenizer VQGAN \cite{esser2021taming} introduces generative training and perceptual loss based on VQVAE, resulting in enhanced perceptual quality. Several studies have employed such tokenizers to implement image codecs. Mao et al. \cite{mao2024extreme} proposed adjusting the codebook size via K-means clustering in a pretrained VQGAN. By discarding and regenerating unimportant tokens via transformers, the bitrate is further reduced \cite{xue2024unifying}. Jia et al. \cite{Jia_2024_CVPR} introduced transform coding in the latent space of VQGAN to achieve impressive performance. 
More recently, the 1-D tokenizer \cite{yu2024imageworth32tokens} improved semantic compression by reducing spatial redundancy; however, its fixed resolution poses challenges for flexible compression applications.
In this paper, we reference the 1-D tokenizer for compact semantic compression in a patch-wise manner, enabling variable resolution. Furthermore, our work is tailored with adaptive detail coding, enhancing fidelity beyond that of the 1-D tokenizer alone.

Our work is related to HybridFlow \cite{lu2024hybridflow}, which combines discrete tokens from a VQGAN tokenizer with continuous features from an MLIC encoder \cite{jiang2023mlic}. However, our approach differs in three key aspects:
(1) By adopting independent VQGAN tokenizer and MLIC encoder, HybridFlow contains non-trivial information  redundancy between the two branches. By contrast, DLF focuses on inter-branch redundancy reduction via cross-branch interaction.
(2) HybridFlow relies on a manually pre-defined masking strategy for token reduction, rather than learning from the content, limiting its adaptability to the diverse images. In comparison, DLF aggregates semantics into 1-D tokens through learning, thereby enhancing flexibility.
(3) Since HybridFlow uses a fixed MLIC encoder, its detail features cannot adapt to generation errors in the tokenizer branch. By contrast, the detail encoder of DLF learns to dynamically correct generation through joint optimization, enhancing final quality.
These innovations help DLF achieve $2.6\times$ higher compression ratio than that of HybridFlow.

\begin{figure*}[t]
    \begin{center}
        \includegraphics[width=1.0\linewidth]{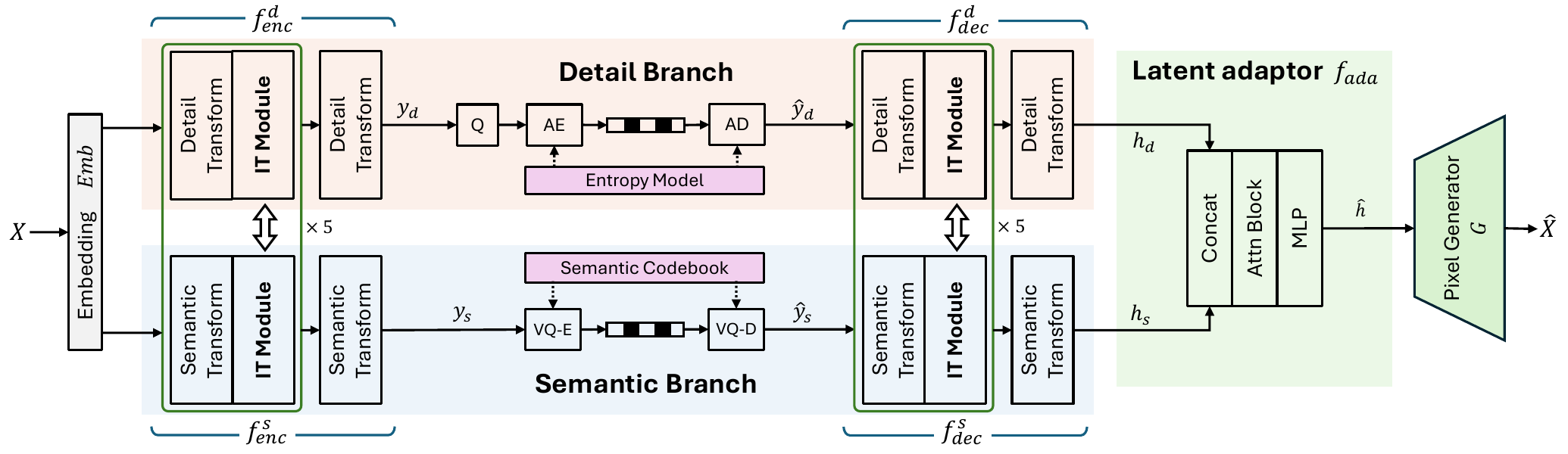}
    \end{center}
    \vspace{-6mm}
    \caption{
        Overview of the proposed DLF framework. 
        Q refers to scalar quantization, AE and AD denote arithmetic encoding and decoding, VQ-E and VQ-D stand for VQ-indices encoding and decoding.
        $f_{enc}^{s}$ and $f_{dec}^{s}$ represent the encoder and decoder parts of the semantic branch, while $f_{enc}^{d}$ and $f_{dec}^{d}$ denote the corresponding parts in detail branch.
        Both branches use multiple Interactive Transform (IT) modules to facilitate learning the complementary information.
    }
    \label{fig:method_1}
    \vspace{-5mm}
\end{figure*}

\subsection{Diffusion-based Extreme Image Compression} 
Diffusion models \cite{rombach2021highresolution, pernias2023wuerstchen} have shown impressive capabilities in generative compression tasks \cite{lei2023textsketchimage, careil2023towards, li2024extremeimagecompressionlatent, li2024diffusionbasedextremeimagecompression, xu2024idempotence, zhang2025stablecodec, xu2025picd, xue2025one}.
Text+Sketch \cite{lei2023textsketchimage} adopts Stable Diffusion \cite{rombach2021highresolution} for image compression, conditioned on text prompts and structure maps.
Despite improved realism, their fidelity remains unsatisfactory.
In contrast, PerCo \cite{careil2023towards} fine-tunes a diffusion model within an image codec, incorporating vector-quantized spatial features and global text descriptions.
It achieves remarkable reconstruction quality at extremely low bitrates below 0.01 bpp.
Recently, DiffEIC \cite{li2024extremeimagecompressionlatent} and RDEIC \cite{li2024diffusionbasedextremeimagecompression} use VAEs to compress the image latent for diffusion guidance to achieve superior performance.
However, despite these guidance techniques, diffusion models still often generate irrelevant content at extremely low bitrates, leading to suboptimal reconstruction fidelity.
Compared with diffusion-based methods, our proposed dual-branch coding framework delivers enhanced fidelity, competitive realism, and significantly faster decoding speed.

\section{Method}

\subsection{Dual-branch Coding Pipeline}
\noindent Unlike existing methods \cite{mao2024extreme, xue2024unifying, Jia_2024_CVPR} that employ the generative tokenizer \cite{esser2021taming} in a single branch, our DLF introduces an innovative dual-branch coding pipeline. 
As illustrated in Fig.~\ref{fig:method_1}, it decomposes the latent representation into semantic and detail components. 
In the semantic branch, DLF clusters high-level semantics into compact tokens by learning a codebook from a large training dataset. 
It then employs a detail branch to capture diverse individual-level details through a large quantization space, complementing the semantic branch.
In addition, DLF incorporates a cross-branch interaction mechanism to minimize redundancy between the semantic and detail branches. 

\vspace{1mm}
\noindent \textbf{Encoding.} 
We first embed the input image $X \in \mathbb{R} ^{3 \times H \times W}$ into $Emb(X) \in \mathbb{R} ^{C \times h \times w}$ through patch embedding, with a patch size of $h = H/16$ and $w = W/16$. 
This embedding is then fed into dual branches to extract the semantic feature $y_s$ and detail feature $y_d$:
\begin{equation}
    y_{s} = f_{enc}^{s}(Emb(X)), \quad y_{d} = f_{enc}^{d}(Emb(X))
\end{equation} 

For the semantic branch, we adopt a 1-D tokenizer \cite{yu2024imageworth32tokens} to learn dataset-level semantic clustering to facilitate efficient low-bitrate encoding.
Specifically, we first partition $Emb(X)$ into $16 \times 16$ windows. Within each window, both the $16 \times 16$ 2-D image tokens and 32 extra 1-D tokens are fed into a vision transformer (ViT). 
Through cascaded attention blocks, these 1-D tokens efficiently aggregate essential semantic information from the image tokens, thereby minimizing spatial redundancy.
Unlike previous methods \cite{xue2024unifying, lu2024hybridflow} that use manually pre-defined masks for token reduction, our approach compresses semantics into fewer tokens through large-scale learning, offering greater flexibility in handling various images.
Consequently, we encode these compact 1-D tokens to yield $y_s \in \mathbb{R} ^{N \times C \times 32}$, where $N=\frac{h \times w}{16 \times 16}$ denotes the number of windows in feature map.

\begin{figure}[t]
    \begin{center}
        \includegraphics[width=1.0\linewidth]{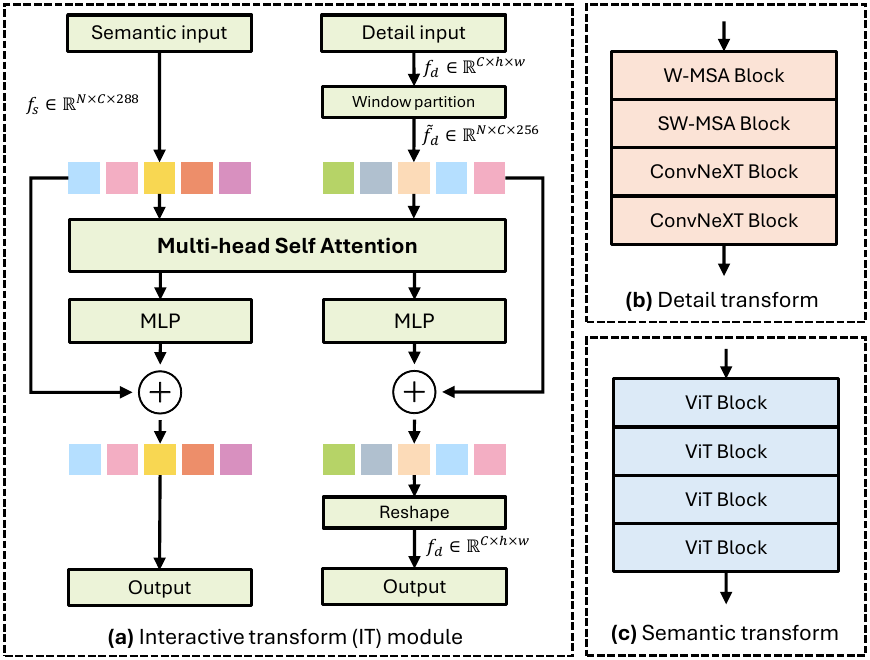}
    \end{center}
    \vspace{-6mm}
    \caption{
        \textbf{(a)} Interactive Transform (IT) module.
        For each window in semantic branch, we have $16 \times 16$ feature tokens and 32 additional tokens, thus the length of the sequence $f_s$ is 288.
        \textbf{(b)} Detail transform block. 
        We employs shifted window based multi-head self-attention (W-MSA and SW-MSA) \cite{Liu_2021_ICCV} and ConvNeXT block \cite{Liu_2022_CVPR} in this module.
        \textbf{(c)} Semantic transform block. 
        We adopts standard ViT module to extract semantics for each window.
    }
    \label{fig:method_2}
    \vspace{-4mm}
\end{figure}

For the detail branch, we use a combination of shifted window attention \cite{Liu_2021_ICCV} and convolution blocks \cite{Liu_2022_CVPR}.
This design allows the model to extract important local details while remaining aware of the global detail distribution.
The detail features will be downsampled to $y_d \in \mathbb{R} ^{C \times \frac{h}{2} \times \frac{w}{2}}$ to reduce the bitrate.
Fig.~\ref{fig:method_2} (b-c) illustrates the structure of the semantic and detail transform blocks.

\begin{figure}[t]
    \centering
    \includegraphics[width=1.0\linewidth]{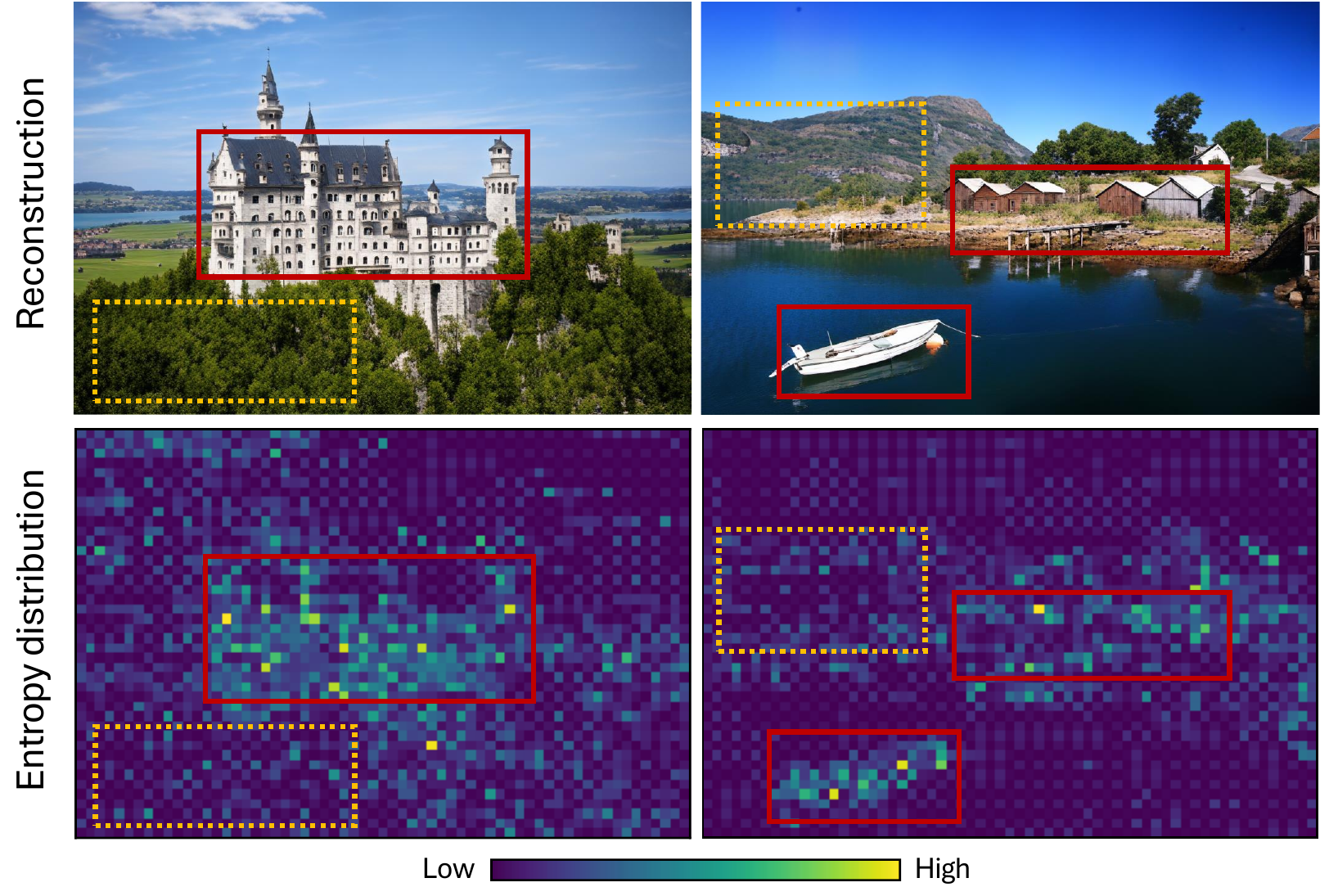}
    \vspace{-6mm}
    \caption{
        Entropy distribution of the quantized detail feature.
        \textcolor{red}{Solid boxes} are details of specific objects, which consume more bits.
        \textcolor{orange}{Dashed boxes} are relatively more common contents, which require fewer bits despite also having high-frequency information.
    }
    \label{fig:method_4}
    \vspace{-6mm}
\end{figure}

\vspace{1mm}
\noindent \textbf{Quantization.}
We learn a codebook to encode the semantic latents $y_s$ through vector quantization (VQ). 
The resulting codebook indices are encoded using fixed-length coding for simplicity.
For detail latents $y_d$, we employ scalar quantization (SQ) and use quadtree partition-based entropy model \cite{Li_2023_CVPR, li2024neural} to estimate their distribution for arithmetic coding.
Compared to VQ, SQ effectively leverages a much larger quantization space \cite{mentzer2023finitescalarquantizationvqvae} to represent various individual-level details. 
In particular, we can easily apply different learnable quantization steps to SQ for spatially varying regions. So the detail branch can adaptively allocate more bits to encode the most critical information.
The quantization process in DLF can be formulated as:
\begin{equation}
    \hat{y}_{s} = VQ(y_s), \quad \hat{y}_{d} = Q(y_d)
\end{equation} 

Fig.~\ref{fig:method_4} presents the entropy distribution of the quantized detail features.
For relatively common content, such as mountains and trees, the semantic branch can effectively encode them because the codebook is enriched with this type of information frequently occurring in the dataset.
Consequently, the detail branch requires very few bits to represent these regions, even if they may contain high-frequency information.
By contrast, relatively unique objects like the outlines of the castle and boat are more challenging to represent with highly clustered tokens, so the detail branch adaptively allocates more bits to encode these objects.
This illustrates how DLF leverages the semantic and detail branches to dynamically encode images, thus achieving both lower bitrate and improved visual fidelity.
\vspace{1mm}
\noindent \textbf{Decoding.}
The quantized semantic and detail latents are decoded by their corresponding branch, producing the intermediate features $h_s \in \mathbb{R} ^{C \times h \times w}$ and $h_d \in \mathbb{R} ^{C \times h \times w}$:
\vspace{-1mm}
\begin{equation}
    h_{s} = f_{dec}^{s}(\hat{y}_{s}), \quad h_{d} = f_{dec}^{d}(\hat{y}_{d})
\end{equation}
%
Subsequently, the intermediate features are fused through a latent adaptor and ultimately converted into the reconstructed image $\hat{X}$ using a pixel generator $G$:
\vspace{-1mm}
\begin{gather}
    \hat{h} = f_{ada}(h_d, h_s) \\[0.25em]
    \hat{X} = G(\hat{h})
\end{gather}
%
To enhance reconstruction quality, we use a pretrained VQGAN decoder as the pixel generator \cite{esser2021taming} and fine-tune it with the entire model.

\subsection{Cross-branch Interactive Design}
The dual-branch design of DLF requires eliminating redundancy between the semantic and detail bottlenecks.
Furthermore, the window-wise encoding in the semantic branch requires inter-window interactions to capture global semantic information.
To meet these requirements, we propose a cross-branch interaction design within DLF, implemented through a stacked Interactive Transform (IT) module.
As depicted in Fig.~\ref{fig:method_2} (a), the detail feature $f_d \in \mathbb{R}^{c \times h \times w}$ is initially partitioned into $\tilde{f_d} \in \mathbb{R} ^{N \times C \times 256}$ using the same strategy as in the semantic branch.
Subsequently, the semantic and detail features are jointly fed into the multi-head self-attention layer to facilitate interaction.
Finally, the processed detail feature is reshaped back into its original form.
Such design offers two significant advantages.
First, the self-attention layer dynamically aggregates the semantic and detail information through joint training, reallocating them to each branch.
This gradual aggregation and reallocation not only mitigate redundancy between branches but also enable detail information to dynamically correct generation errors in the semantic branch.
Second, the interaction provides the semantic branch with cross-window perception. The detail feature contributes global information through this interaction, extending the semantic branch's receptive field beyond individual windows and thereby enhancing semantic capturing capability.

\begin{figure}[t]
    \begin{center}
        \includegraphics[width=1.0\linewidth]{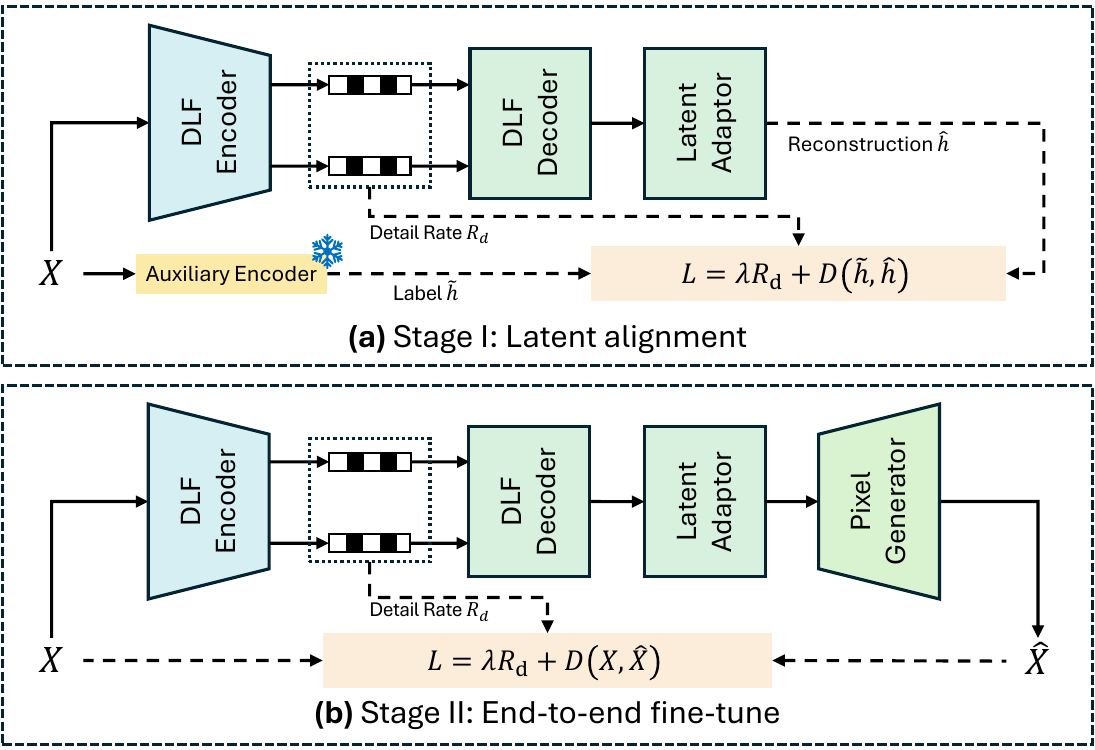}
    \end{center}
    \vspace{-6mm}
        \caption{
        Progressive training of DLF. 
        \textbf{(a)} In stage I, we train DLF codec and latent adaptor in latent domain.
        In this stage, the pretrained generative tokenizer provides label for supervision.
        \textbf{(b)} In stage II, we fine-tune the entire DLF model in pixel domain.
    }
    \label{fig:method_3}
    \vspace{-4mm}
\end{figure}

\begin{figure*}[t]
    \begin{center}
        \includegraphics[width=1.0\linewidth]{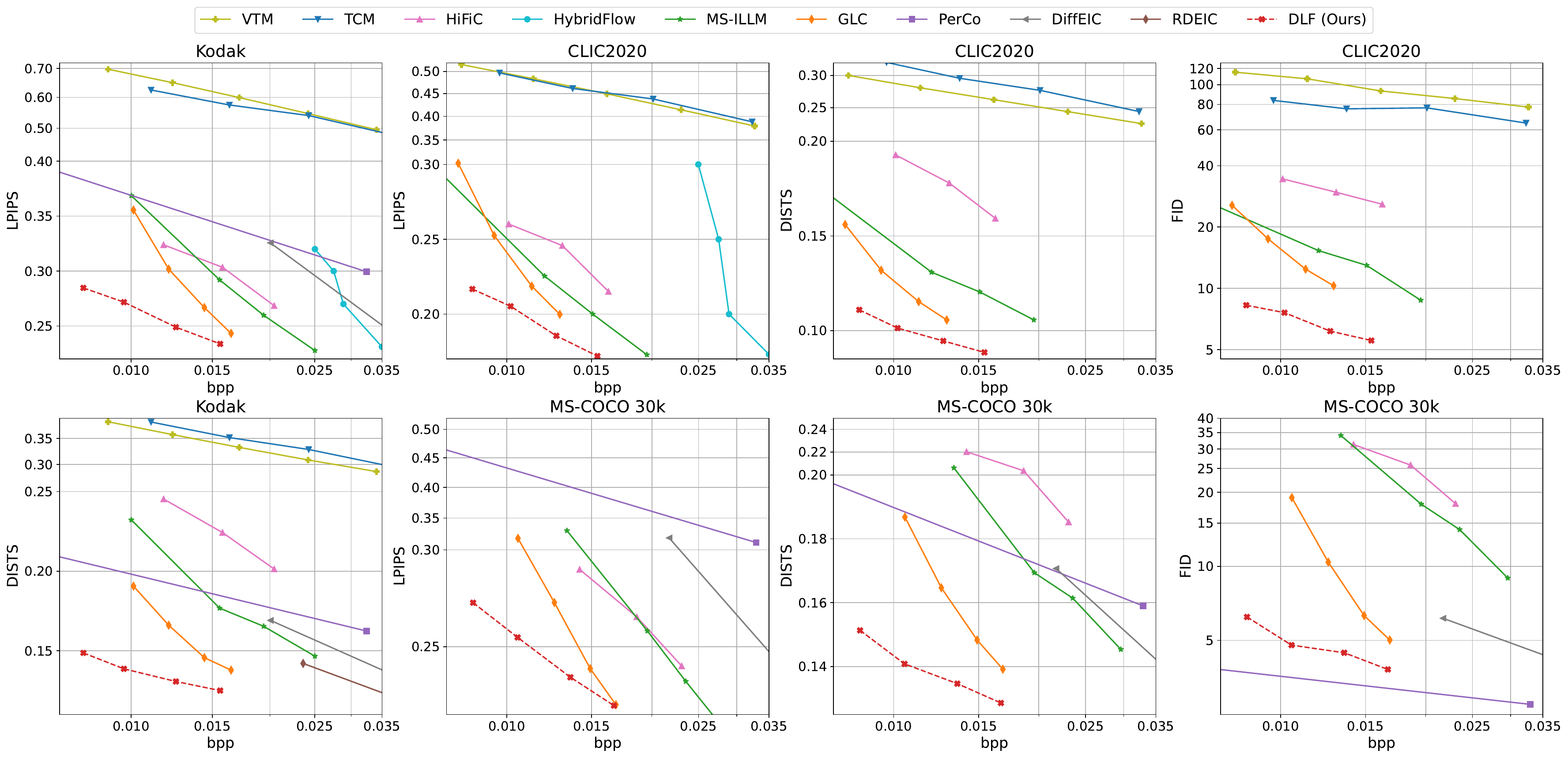}
    \end{center}
    \vspace{-5mm}
    \caption{
        Rate-distortion curves on the Kodak, the CLIC2020 and the MS-COCO 30K datasets. See supplementary for larger version.
    }
    \label{fig:exp_1}
\end{figure*}

\begin{figure*}[h]
    \begin{center}
        \includegraphics[width=1.0\linewidth]{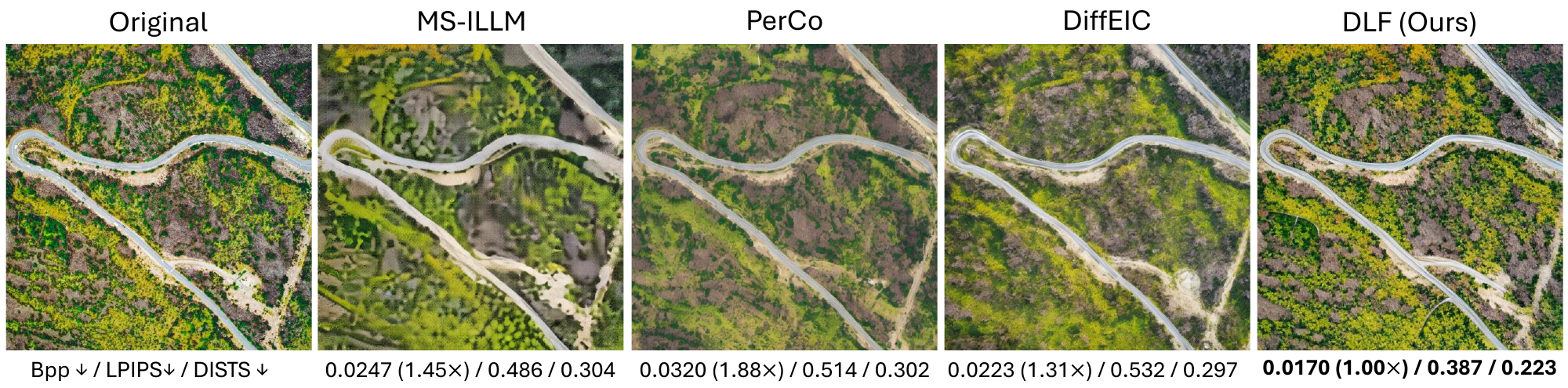}
    \end{center}
    \vspace{-5mm}
    \caption{
        Qualitative examples on the CLIC2020 ($768\times768$) dataset. More comparisons are in supplementary.  
    }
    \label{fig:exp_2_clic}
    \vspace{-1mm}
\end{figure*}

\subsection{Two-stage Training Strategy}
To ensure compatibility with the pixel generator, we adopt a two-stage progressive training inspired by previous works \cite{yu2024imageworth32tokens, Jia_2024_CVPR}.
As shown in Fig. \ref{fig:method_3}, this strategy involves two stages: latent alignment and end-to-end fine-tune.
In the latent alignment stage, we apply a rate-distortion loss in the latent space for alignment. Given the pretrained pixel generator, we use its corresponding encoder \cite{esser2021taming} (i.e. the auxiliary encoder in Fig.\ref{fig:method_3}) to generate  $\tilde{h}$ to supervise the latent reconstruction $\hat{h}$. The bitrate is calculated based on the distribution of the quantized detail latents. After the first stage, we further fine-tune the entire model using the pixel-space generative loss \cite{esser2021taming}. During the training process, the detail branch will learn the spatial adaptive compression and allocate more bits to the most important regions for fidelity. Further details on the training and block design are provided in the supplementary material.

\section{Experiments}

\subsection{Implementation Details}
\noindent \textbf{Training details.} 
DLF is trained on the subset of the OpenImages train set \cite{openimages}. During stage I, we train the model on randomly cropped $256 \times 256$ patches with a batch size of 8, using a fixed bitrate weight $\lambda$ of $24.0$. In this stage, we incorporate the pretrained 1-D tokenizer \cite{yu2024imageworth32tokens} and freeze its weights. This offers a strong initialization to leverage its semantic prior, facilitating faster convergence. For stage II, we fine-tune the model using randomly cropped $512 \times 512$ patches with a batch size of 4. Here $\lambda$ is set to $\{5.8, 8.5, 16.0, 28.0\}$ to achieve different bitrates. 

\begin{table}[t]
\begin{center}
\caption{
    BD-Rate of generative codecs on the Kodak and CLIC2020 (full resolution) dataset. Anchor: MS-ILLM \cite{muckley2023improving}.
    %
}
\vspace{-2mm}
\label{tab:bd1}
\resizebox{0.43\textwidth}{!}{
\begin{tabular}{l|cccc}
\hline
\multirow{2}{*}{Method}             & \multicolumn{2}{c}{Kodak}             & \multicolumn{2}{c}{CLIC2020}        \\
                                    & LPIPS            & DISTS              & LPIPS            & DISTS            \\ \hline
PerCo \cite{careil2023towards}      & 101.74\%          & -4.02\%           & --                & --                \\
DiffEIC \cite{li2024extremeimagecompressionlatent}      & 66.05\%           & 14.67\%           & --                & --                \\
HybridFlow \cite{lu2024hybridflow}  & 65.30\%           & --                & 145.93\%          & --                \\
HiFiC \cite{mentzer2020high}        & 8.74\%            & 85.14\%           & 27.83\%           & 110.09\%          \\
MS-ILLM \cite{muckley2023improving} & 0.00\%            & 0.00\%            & 0.00\%            & 0.00\%            \\
GLC \cite{Jia_2024_CVPR}            & -17.24\%          & -33.41\%          & -3.59\%           & -19.45\%          \\
DLF                                 & \textbf{-43.05\%} & \textbf{-67.82\%} & \textbf{-27.93\%} & \textbf{-53.55\%} \\ \hline
\end{tabular}
}
\end{center}
\vspace{-5mm}
\end{table}

\begin{figure*}[t]
    \begin{center}
        \includegraphics[width=1.0\linewidth]{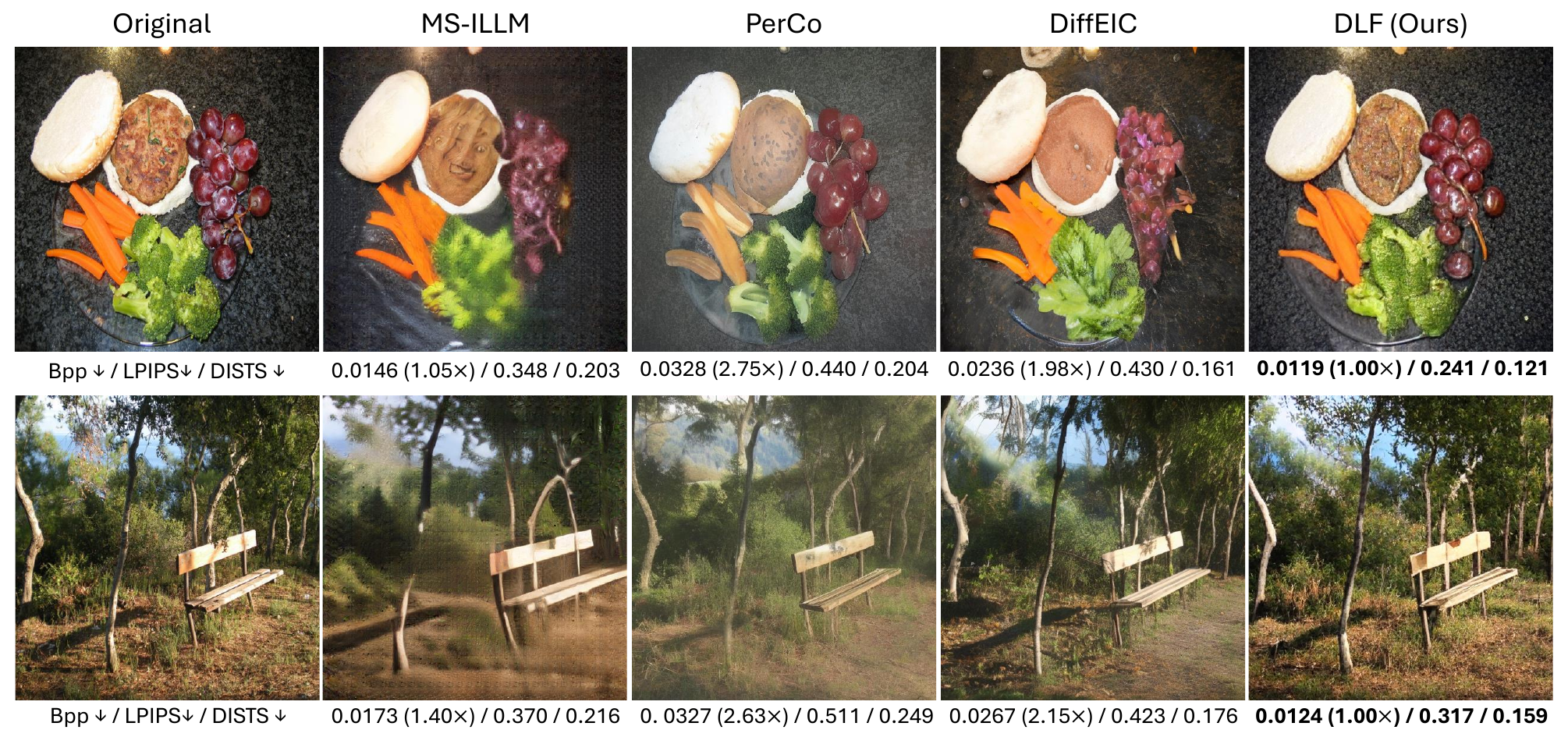}
    \end{center}
    \vspace{-3mm}
    \caption{
        Qualitative examples on the MS-COCO 30K dataset. More comparisons are in supplementary.  
        %
        %
    }
    \label{fig:exp_2_coco}
    \vspace{-1mm}
\end{figure*}

\begin{figure}[h!]
    \begin{center}
        \includegraphics[width=1.0\linewidth]{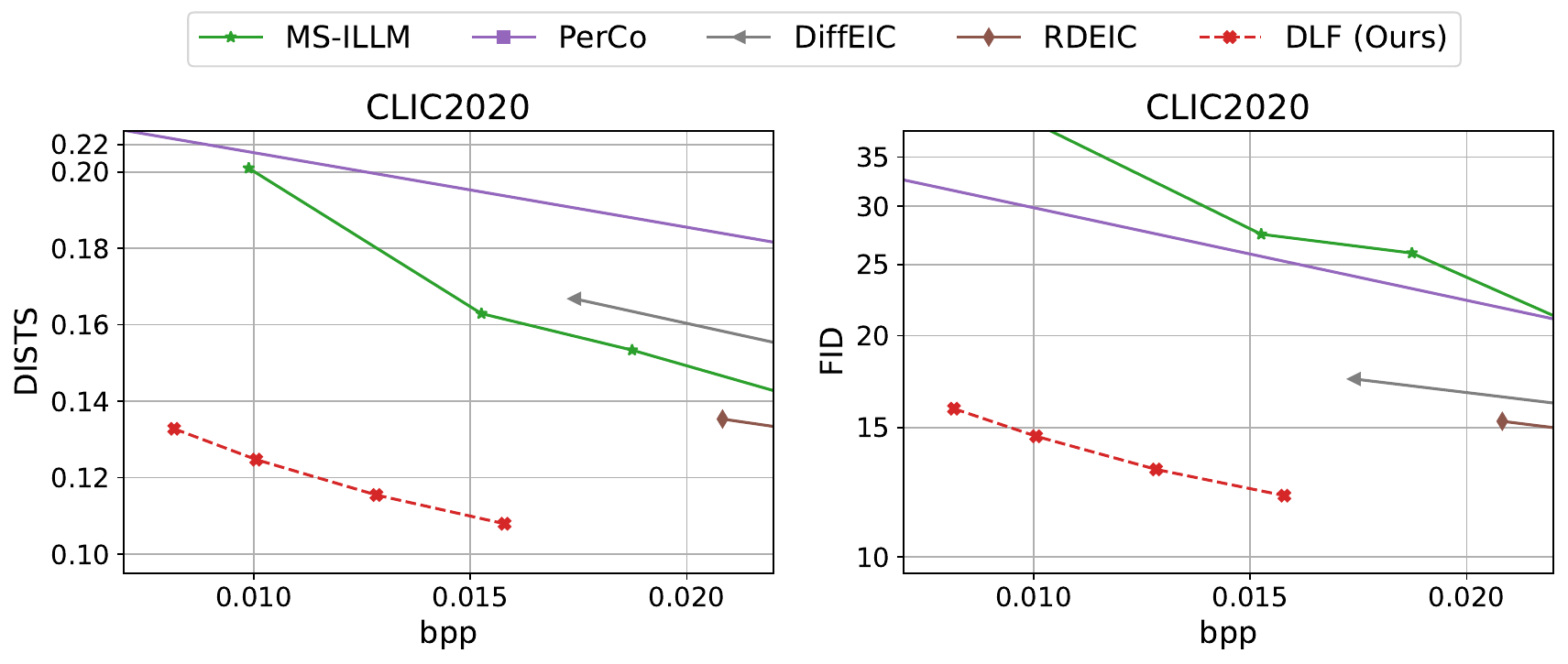}
    \end{center}
    \vspace{-3mm}
    \caption{
        Rate-distortion curves on the CLIC2020 at $768\times768$.
    }
    \label{fig:exp_1_clic768}
    \vspace{-3mm}
\end{figure}

\vspace{1mm}
\noindent \textbf{Evaluation dataset.} 
We evaluate DLF on the widely used Kodak \cite{kodak}, CLIC2020 test \cite{CLIC2020}, and MS-COCO 30K datasets \cite{lin2014microsoft}. For the Kodak and CLIC2020 datasets, we primarily focus on full-resolution images. We also evaluate cropped CLIC2020 images following \cite{yang2024lossy, li2024diffusionbasedextremeimagecompression}, where the short side is resized to 768 pixels and a $768\times768$ center crop is applied. We follow \cite{careil2023towards} to test MS-COCO 30K dataset at a resolution of $512 \times 512$. Results on the DIV2K dataset \cite{Agustsson_2017_CVPR_Workshops} are included in the supplementary material.
\vspace{1mm}
\noindent \textbf{Evaluation metrics.} 
We assess reconstruction fidelity using the reference perceptual metrics LPIPS \cite{Zhang_2018_CVPR} and DISTS \cite{ding2020image}, while generation realism is evaluated with the no-reference perceptual metric FID \cite{heusel2017gans}. The encoded bitrate is calculated in bits per pixel (bpp). It is worth noting that FID is not measured on the Kodak dataset, as its small size (24 images) does not allow for reliable FID evaluation \cite{careil2023towards}. Results for PSNR and MS-SSIM \cite{wang2003multiscale} are provided in our supplementary material.
\vspace{1mm}
\noindent \textbf{Comparison methods.} 
We compare our DLF with the traditional codec VVC \cite{bross2021overview}, the MSE-optimized neural codec TCM \cite{Liu_2023_CVPR}, and generative codecs.
They include the VAE-based HiFiC \cite{mentzer2020high}, MS-ILLM \cite{muckley2023improving}, the tokenizer-based GLC \cite{Jia_2024_CVPR}, the dual-branch-based HybridFlow \cite{lu2024hybridflow} and the diffusion-based PerCo \cite{körber2024percosdopenperceptual}, DiffEIC \cite{li2024extremeimagecompressionlatent} and RDEIC \cite{li2024diffusionbasedextremeimagecompression}. We evaluate generative methods exclusively on the MS-COCO 30K dataset, as it is primarily used for assessing generation quality. Diffusion-based methods are evaluated on the cropped CLIC2020 dataset instead of full resolution due to their limitations in high-resolution compression.

\subsection{Main Results}
\noindent \textbf{Quantitative Evaluation.}
In Fig.~\ref{fig:exp_1}, we present the rate-distortion curves across various datasets. For fidelity metrics, DLF achieves superior performance than all other methods, yielding considerably lower (better) LPIPS and DISTS scores at the same bitrates. We also compute the BD-Rate \cite{bjontegaard2001calculation} on LPIPS and DISTS in Table~\ref{tab:bd1}. Using MS-ILLM as anchor, DLF achieves bitrate savings of 43.05\% for LPIPS and 67.82\% for DISTS on the Kodak dataset, significantly outperforming the recent tokenizer-based GLC and diffusion-based DiffEIC. Additionally, our compression ratio exceeds HybridFlow by more than $2.6\times$, demonstrating a substantial improvement in dual-branch compression. Regarding generation realism measured by FID, our method outperforms all non-diffusion-based approaches on the MS-COCO 30K dataset while showing competitive results against advanced diffusion-based method PerCo. Surprisingly,  on the cropped CLIC2020 dataset at $768\times768$ resolution, DLF surpasses PerCo in terms of FID by a large margin, underscoring its superiority in preserving realism on high-quality datasets. As shown in Fig.~\ref{fig:exp_1_clic768}, DLF achieves the best rate-distortion performance for both FID and DISTS on cropped CLIC2020. These results demonstrate the effectiveness of DLF, highlighting the great potential of our interactive dual-branch coding design.
\noindent \textbf{Qualitative Evaluation.}
In Fig.~\ref{fig:exp_2_clic}, we present a visual comparison from the CLIC2020 test set at a resolution of $768\times768$. The VAE-based MS-ILLM struggle to generate realistic details at a low bitrate, resulting in blurred details and additional artifacts. While diffusion-based methods PerCo and DiffEIC improve realism, they introduce noticeable content distortions (e.g., in vegetation). Additional examples from the MS-COCO 30K dataset (Fig. \ref{fig:exp_2_coco}) show similar trends: MS-ILLM remains blurry, and PerCo and DiffEIC fail to maintain fidelity (e.g., the distorted grape). Compared to these methods, DLF reconstructs visually appealing images with fewer artifacts, accurate colors, and finer details, even at lower bitrates. More visual results are available in the supplementary material.

\begin{table}[t]
\begin{center}
\caption{
    BD-Rate of different model variants. We use the setting \textit{w/ SQ detail} as the anchor, which is finally adopted by DLF.
}
\vspace{-0mm}
\label{table:bd2}
\resizebox{0.42\textwidth}{!}{
\begin{tabular}{l|cccc}
\hline 
\multirow{2}{*}{Model variants} & \multicolumn{2}{c}{Kodak}  & \multicolumn{2}{c}{CLIC2020} \\
                                & LPIPS        & DISTS       & LPIPS         & DISTS        \\ \hline
w/o detail                      & 17.5\%       & 20.2\%      & 47.9\%        & 47.6\%       \\
w/o interactive                 & 64.1\%       & 73.6\%      & 68.8\%        & 61.8\%       \\
w/ VQ detail                    & 18.3\%       & 40.7\%      & 27.3\%        & 58.1\%       \\
w/ SQ detail (\textbf{DLF})     & 0.0\%        & 0.0\%       & 0.0\%         & 0.0\%        \\ \hline
\end{tabular}
}
\end{center}
\vspace{-5mm}
\end{table}

\subsection{Ablation Study}
\noindent In this section, we conduct ablation studies to evaluate the impact of the dual-branch coding architecture, the cross-branch interactive design, and the scalar quantization (SQ) based detail coding design.
All model variants are evaluated using the Kodak and CLIC2020 datasets.
BD-Rate results are presented in Table~\ref{table:bd2}.
\vspace{1mm}
\noindent \textbf{Dual-branch coding architecture.}
We remove the entire detail branch, including all IT modules (denoted as \textit{w/o detail}), and compress the image using only the semantic branch.
In this variant, the bitrate is adjusted solely by varying the number of VQ tokens.
As shown in Table~\ref{table:bd2}, removing detail branch brings a loss in BD-Rate over 47\% on the CLIC2020 dataset, as measured by LPIPS and DISTS.
These findings validate that the dual-branch coding is superior than single-branch semantic coding.
\vspace{1mm}
\noindent \textbf{Cross-branch Interactive Design.}
In this variant, we remove all IT modules, eliminating interactions between the two branches (denoted as \textit{w/o interactive}).
This setting results in the most severe performance drop across all metrics and datasets, with bitrate loss exceeding 60\%.
It indicates that using two parallel but independent branches leads to substantial redundancy between the two bottlenecks, underscoring the crucial role of our interactive design.
\vspace{1mm}
\noindent \textbf{SQ-based Detail Quantization.}
In this variant, we replace the scalar quantization in the detail branch with vanilla vector quantization (denoted as \textit{w/ VQ detail}).
Huffman coding is applied to encode VQ indices, and the bitrate is adjusted via the codebook size.
It also shows a non-trivial performance drop, indicating that the limited quantization space of the VQ codebook is insufficient for capturing various individual-level details, thus highlighting the effectiveness of our spatially adaptive SQ design.
Furthermore, the dual-VQ bottleneck does not outperform the single branch solution, indicating that our model requires a hybrid quantization strategy of VQ-SQ to achieve the best performance.
\subsection{Complexity Analysis}
We compare the average coding time (with standard deviation) in Table~\ref{table:comp1}. Compared to MS-ILLM, DLF incorporates a larger model and thus has a relatively slower coding speed. However, the larger model enables stronger generative capabilities, which form the basis of high-quality reconstruction at extremely low bitrates of 0.01 bpp. Without sufficient generative capabilities, MS-ILLM produces noticeably blurry reconstructions.
While compared to recent extreme image codecs that incorporate large diffusion models (e.g., PerCo and DiffEIC), DLF offers a much faster decoding speed than those methods, along with improved coding performance—representing considerable potential in adopting large tokenizer models for extreme image compression.

\begin{table}[t]
\begin{center}
\caption{
    Complexity analysis and BD-Rate evaluation (Metric: DISTS, anchor: MS-ILLM) on the Kodak dataset.
    The tests are conducted on an NVIDIA A100 GPU.
}
\vspace{-0mm}
\label{table:comp1}
\resizebox{0.43\textwidth}{!}{
\begin{tabular}{l|ccc}
\hline
Model                                     & Enc. Time (s)          & Dec. Time (s)        & BD-Rate   \\ \hline
MS-ILLM \cite{muckley2023improving}       & 0.064 ± 0.010          & 0.070 ± 0.011        & 0.00\%     \\
PerCo \cite{careil2023towards}            & 0.461 ± 0.017          & 2.443 ± 0.011        & -4.02\%     \\
DiffEIC \cite{li2024extremeimagecompressionlatent} & 0.152 ± 0.014 & 4.093 ± 0.042        & 14.67\%      \\
DLF                                       & 0.178 ± 0.015          & 0.252 ± 0.014        & -67.82\%     \\ \hline
\end{tabular}
}
\end{center}
\vspace{-5mm}
\end{table}

\section{Conclusion}
\noindent We introduce a novel neural image codec named Dual-generative Latent Fusion (DLF), targeting at extremely low bitrate.
DLF improves latent coding efficiency via a dual-branch architecture, which decomposes the latent into semantic and detail parts, applying suitable compression methods according to their characteristics.
Furthermore, our cross-branch interactive design minimizes information redundancy between the two bottlenecks.
Experiments validate that DLF achieves SOTA reconstruction quality at bitrates below 0.03 bpp.
Additionally, compared with diffusion-based models, DLF sustains high decoding speed while effectively supporting high-resolution coding.
\noindent \textbf{Limitations.}
Currently, the actual coding speed of our DLF does not meet real-time requirements. In future work, we aim to optimize the model design and enhance the speed.

{
    \small
    \bibliographystyle{ieeenat_fullname}
    \bibliography{main}
}

\clearpage

\appendix

\section{Training Details}
This section outlines the detailed training strategy. We train our model using the Open Images v4 training dataset \cite{openimages}. Due to the storage limitations, we randomly sample 400,000 images from this dataset for training.

\subsection{Stage 1: Latent Alignment}
To ensure that the detail branch captures the most significant contents while discarding minor ones, we impose a rate constraint on the detail features. Consequently, we design the latent domain rate-distortion loss as follows:
\begin{equation}
    \mathcal{L}_{\text{Stage-1}} = ||\hat{h}-\tilde{h}||^2_2 + \lambda \cdot \mathcal{R}(\hat{y}_d)
\end{equation}
where $\hat{h}$ is the decoded latent generated by the latent adaptor, and the supervision latent $\tilde{h}$ is obtained from a pretrained auxiliary encoder (i.e., the VQGAN encoder \cite{esser2021taming}). Additionally, $\mathcal{R}(\hat{y}_d)$ denotes the bitrate of the quantized detail latent $\hat{y}_d$, estimated using the quadtree-partition-based spatial context module \cite{Li_2023_CVPR}. The weight $\lambda$ controls the trade-off between rate and distortion items.

At the start of this stage, we load the pretrained weights of the 1-D tokenizer \cite{yu2024imageworth32tokens} into the semantic branch. These weights are fixed during this stage to maintain their initialization benefits. This approach provides a strong initialization, accelerating the training process and aiding faster convergence. The semantic codebook size is set to 4096.

To prevent the detail branch from discarding excessive information at the training beginning, we employ a multistage $\lambda$ strategy. Specifically, we initiate the training with a $\lambda$ value of 0.001 for the first 10,000 steps. Subsequently, the $\lambda$ value increases gradually from 2.0 to 24.0 over 90,000 steps. Finally, we maintain $\lambda=24.0$ for the remaining 400,000 steps. We apply this strategy to all rate points during this stage, serving as an initialization for the subsequent stage. During this stage, we use randomly cropped $256 \times 256$ images with a batch size of 16, setting the learning rate to $4.0 \times 10^{-5}$ with the Adam optimizer \cite{KingBa15}.
To prevent the detail branch from discarding excessive information at the training beginning, we employ a multistage $\lambda$ strategy. Specifically, we initiate the training with a $\lambda$ value of 0.001 for the first 10,000 steps. Subsequently, the $\lambda$ value increases gradually from 2.0 to 24.0 over 90,000 steps. Finally, we maintain $\lambda=24.0$ for the remaining 400,000 steps. We apply this strategy to all rate points during this stage, serving as an initialization for the subsequent stage. During this stage, we use randomly cropped $256 \times 256$ images with a batch size of 16, setting the learning rate to $4.0 \times 10^{-5}$ with the Adam optimizer \cite{KingBa15}.

\subsection{Stage 2: End-to-end Fine-tune}
In this stage, we fine-tune the entire model, including all parameters in both branches, the latent adaptor, and the pixel generator. To achieve superior reconstruction quality, we employ the pixel-domain rate-distortion loss:
\begin{gather}
    \mathcal{L}_{\text{Stage-2}} = \mathcal{L}_{\text{pixel}} + \mathcal{L}_{\text{codebook}} + \lambda \cdot \mathcal{R}(\hat{y}_d)
\end{gather}
Here, the pixel-domain distortion loss $\mathcal{L}_{\text{pixel}}$ and the codebook loss $\mathcal{L}_{\text{codebook}}$ in the semantic branch are defined as:
\begin{gather}
    \mathcal{L}_{\text{pixel}} = ||x-\hat{x}|| + \mathcal{L}_{\text{LPIPS}}(x,\hat{x}) + \lambda_{\text{adv}} \cdot \mathcal{L}_{\text{adv}}(x,\hat{x}) \\[0.2em]
    \mathcal{L}_{\text{codebook}} = ||\text{sg}(y_s) - \hat{y_s}|| + \beta \cdot ||\text{sg}(\hat{y_s}) - y_s|| 
\end{gather}
The term $\mathcal{L}_{\text{adv}}$ corresponds to the adaptive adversarial loss \cite{esser2021taming}, weighted by $\lambda_{\text{adv}} = 0.8$. The function $\text{sg}(\cdot)$ denotes the stop-gradient operation, and the weight $\beta$ is set to 0.25.

In the first 400,000 steps of this stage, we use randomly cropped $256 \times 256$ images. Subsequently, we switch to $512 \times 512$ images for the next 400,000 steps of training. The enlarged training image size enables the model to be aware of cross-window interactions. In this stage, the $\lambda$ values are set to \{5.8, 8.5, 16.0, 28.0\} for different bitrates, and the model is trained with a learning rate of $2.0 \times 10^{-5}$ and a batch size of 8.

\begin{figure*}[t]
    \centering
    \includegraphics[width=1.0\linewidth]{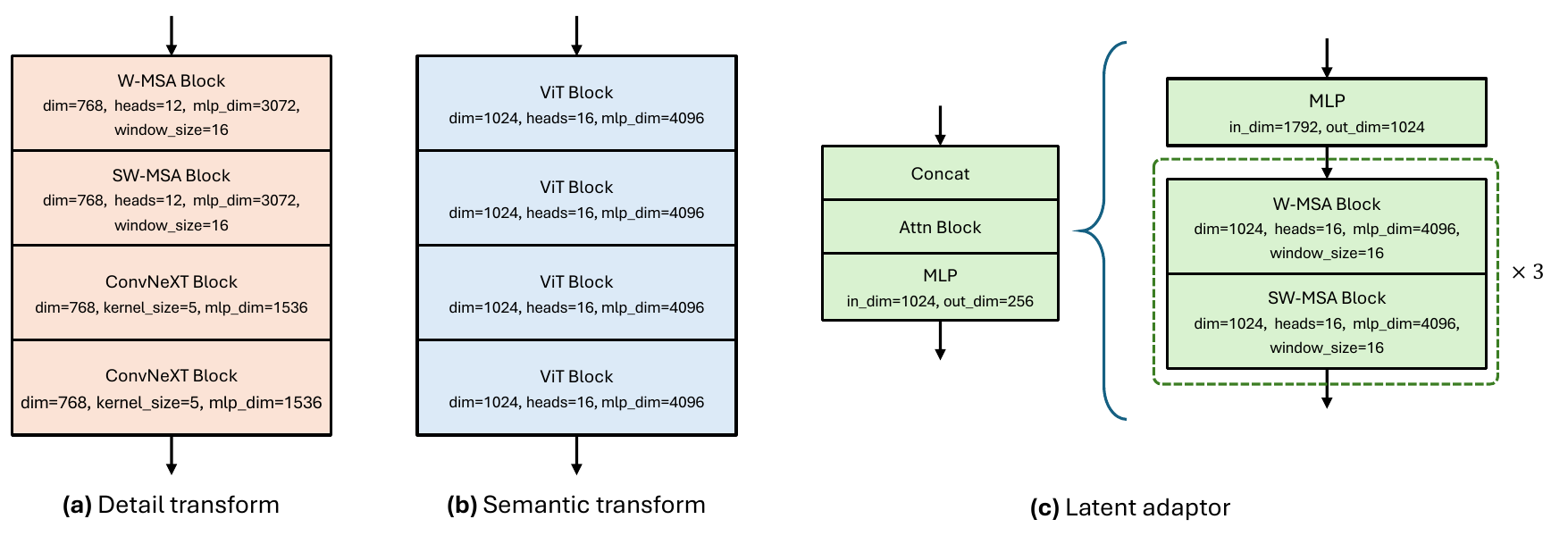}
    \vspace{-8mm}
    \caption{Hyper-parameter settings for the detail and semantic transform blocks, as well as the latent adaptor.}
    \label{fig:arch_1}
    \vspace{-3mm}
\end{figure*}

\begin{figure}
    \centering
    \includegraphics[width=1.0\linewidth]{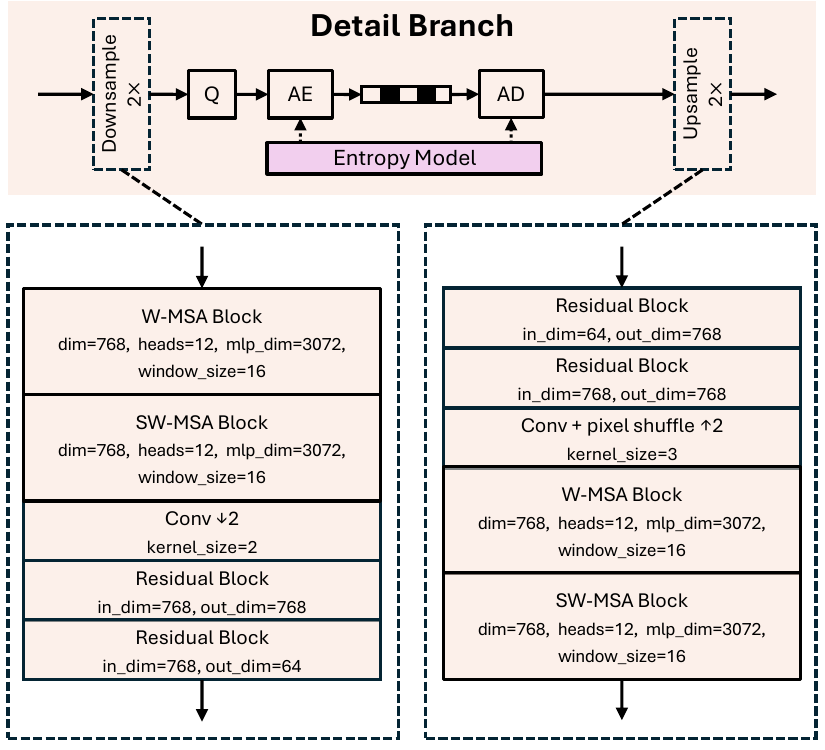}
    \vspace{-5mm}
    \caption{The structures and hyper-parameters of the downsample and upsample modules within the detail branch.}
    \label{fig:arch_2}
    \vspace{-3mm}
\end{figure}

\begin{figure}[t]
    \centering
    \includegraphics[width=1.0\linewidth]{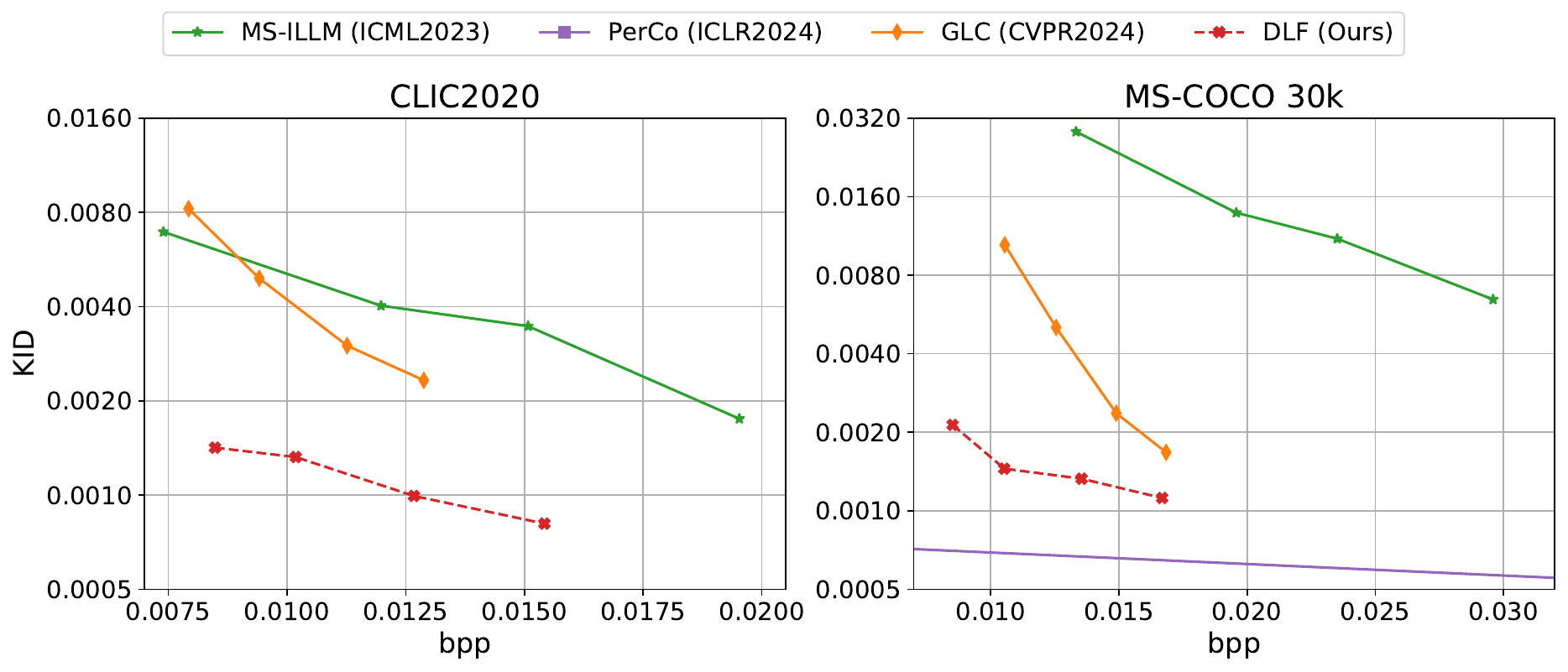}
    \vspace{-6mm}
    \caption{Rate-KID curves on the CLIC2020 and the MS-COCO30k datasets.}
    \label{fig:exp_KID}
    \vspace{-6mm}
\end{figure}

\begin{figure}
    \centering
    \includegraphics[width=0.8\linewidth]{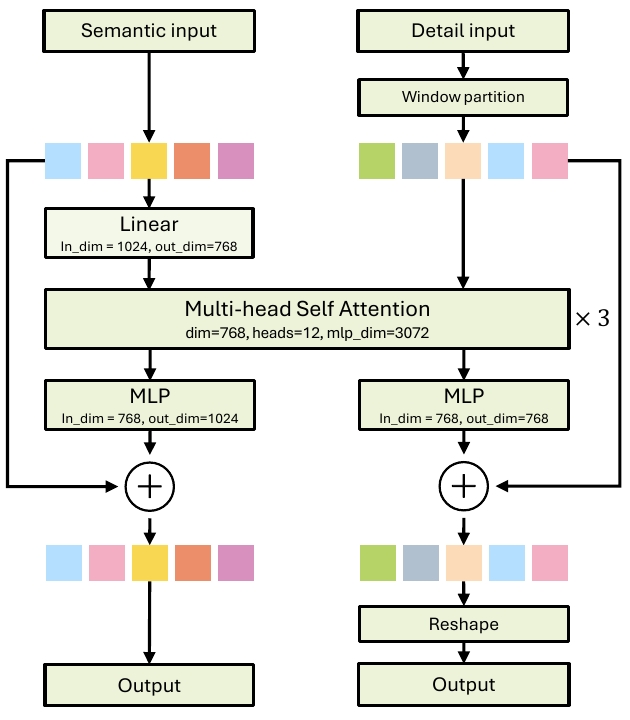}
    \vspace{-3mm}
    \caption{Hyper-parameter settings for the Interactive Transform (IT) block.}
    \label{fig:arch_3}
    \vspace{-3mm}
\end{figure}

\section{Model Architectures}
We illustrate the model architecture with the detailed hyper-parameters. Fig. \ref{fig:arch_1} shows the architecture of the semantic transform, the detail transform and the latent adaptor. Fig. \ref{fig:arch_2} illustrates the downsample module before the scalar quantization and the upsample module after the quantization within the detail branch. Lastly, Fig. \ref{fig:arch_3} depicts the design of the Interactive Transform (IT) module.

\begin{figure*}[t]
    \centering
    \includegraphics[width=0.9\linewidth]{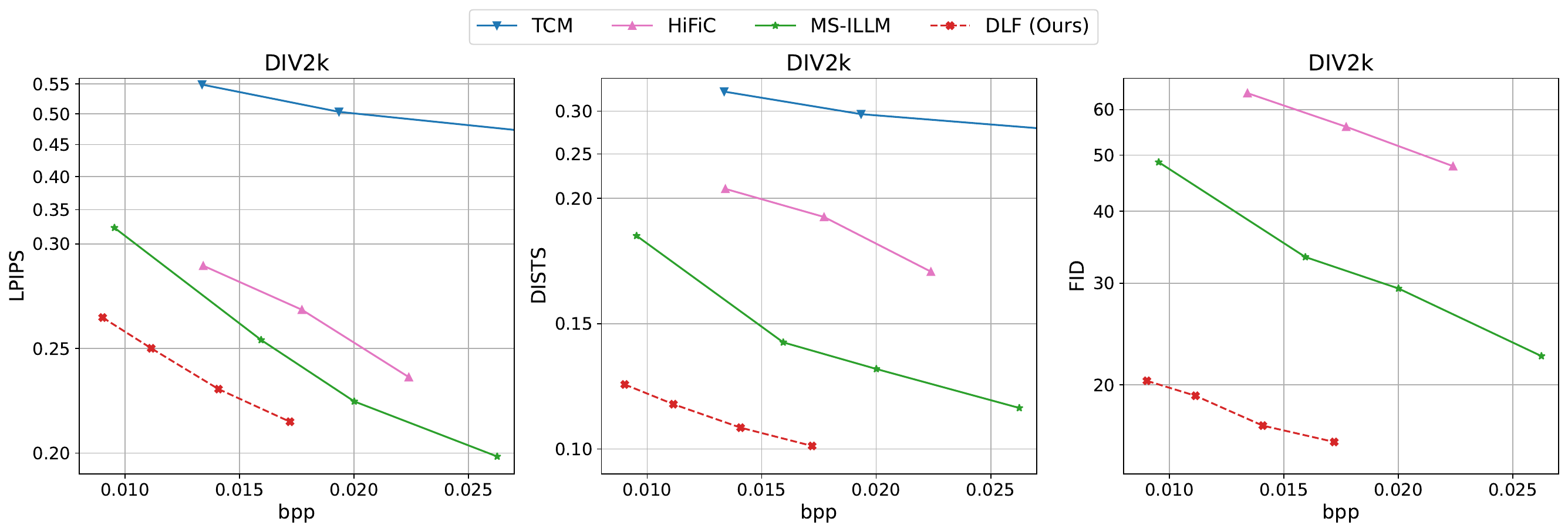}
    \vspace{-2mm}
    \caption{Comparison of methods on the DIV2K dataset \cite{Agustsson_2017_CVPR_Workshops}.}
    \label{fig:exp_DIV2k}
\end{figure*}

\begin{figure*}[t]
    \centering
    \includegraphics[width=0.9\linewidth]{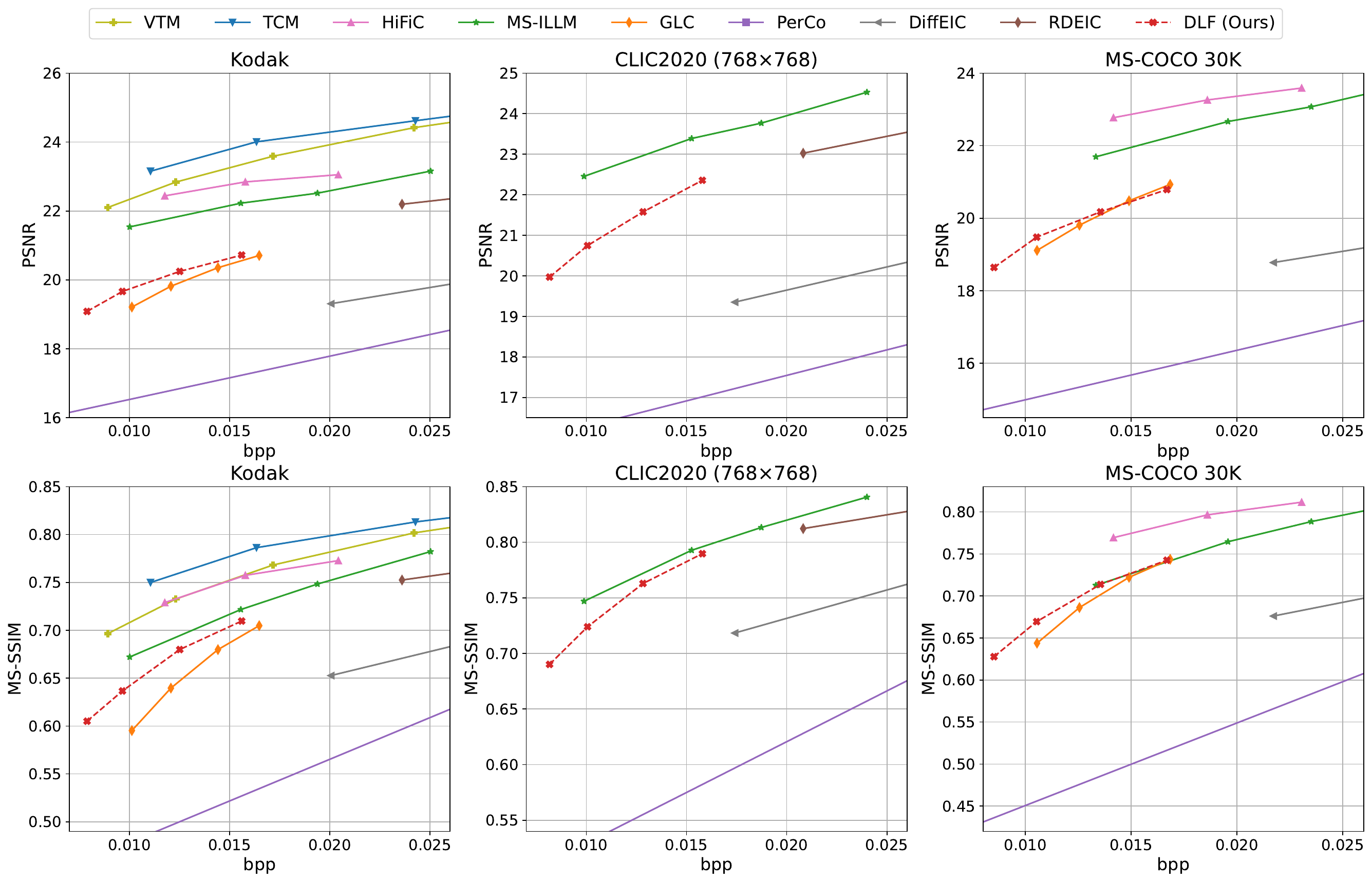}
    \vspace{-2mm}
    \caption{Comparison of methods measured by PSNR and MS-SSIM.}
    \label{fig:exp_PSNR}
\end{figure*}

\section{Experiments}
\subsection{Evaluation details}
\noindent \textbf{Evaluation of third-party models.} We evaluate TCM \cite{Liu_2023_CVPR}, HiFiC \cite{mentzer2020high}, and MS-ILLM \cite{muckley2023improving} by utilizing their official codes and fine-tuning their pretrained models (at the lowest bitrate) to achieve extremely low bitrate ranges. For DiffEIC \cite{li2024extremeimagecompressionlatent}, we employ their released models for evaluation. In the case of PerCo \cite{careil2023towards}, we rely on a third-party implementation \cite{körber2024percosdopenperceptual} for evaluation due to the absence of official code. Similarly, GLC \cite{Jia_2024_CVPR} and RDEIC \cite{li2024diffusionbasedextremeimagecompression} also lack public code. For GLC, we obtain the evaluation results through the personal communication with the authors. For RDEIC and HybridFlow \cite{lu2024hybridflow}, we derive the number from their published papers, since there are no code or data available.
\vspace{1mm}
\noindent \textbf{Measurement of FID and KID.} For the CLIC2020 test dataset \cite{CLIC2020} with full resolution, the FID \cite{heusel2017gans} and KID \cite{binkowski2018demystifying} metric is evaluated by splitting the images into overlapped $256 \times 256$ patches, following the method in HiFiC \cite{mentzer2020high}. This setting is also applied to the CLIC2020 test dataset with $768 \times 768$ resolution, in accordance with the condition in DiffEIC \cite{li2024extremeimagecompressionlatent}. For the MS-COCO 30K dataset \cite{lin2014microsoft}, we directly evaluate the FID and KID on $512 \times 512$ resolution.

\subsection{Quantitative Results}
\begin{figure*}
    \centering
    \includegraphics[width=1.0\linewidth]{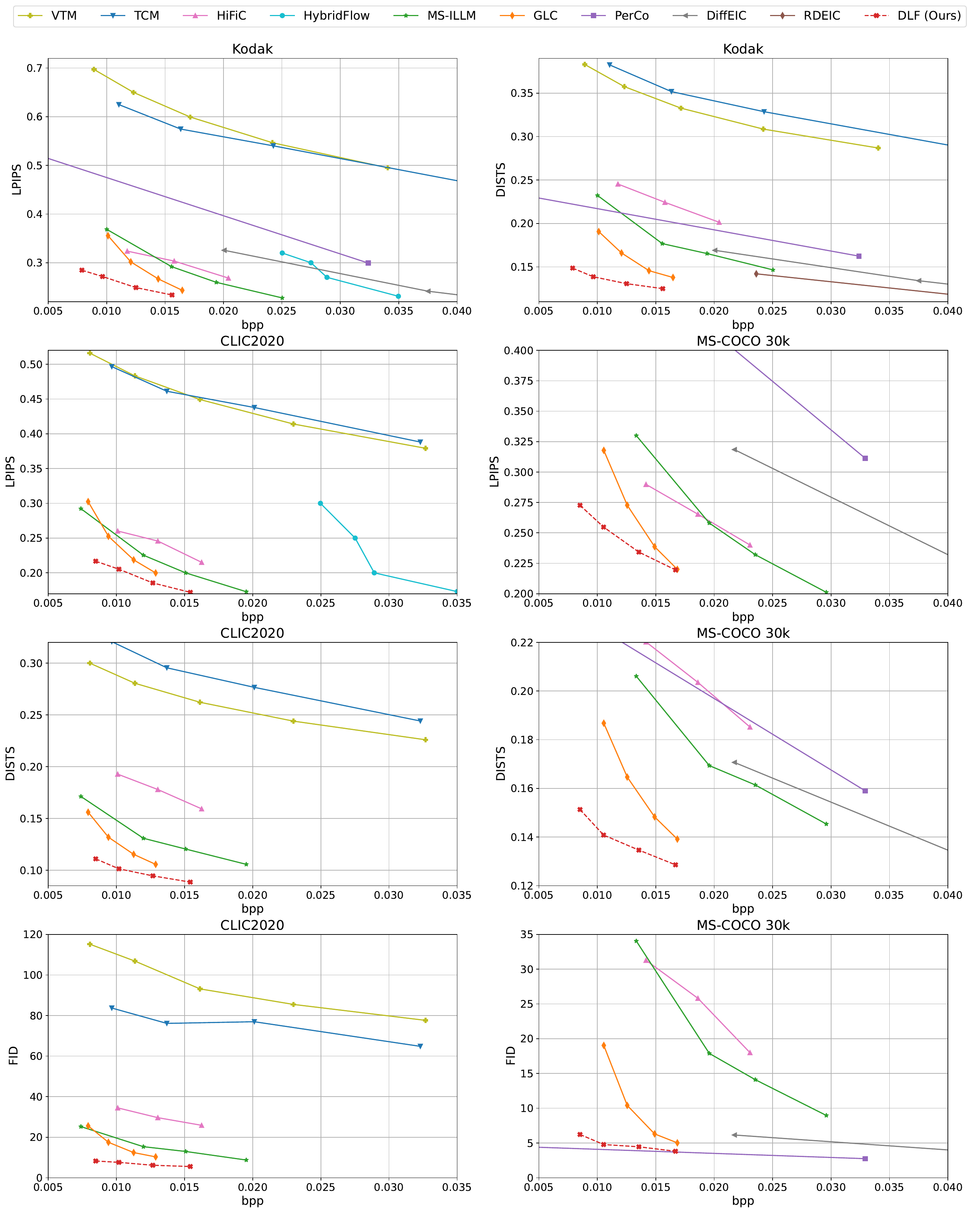}
    \vspace{-3mm}
    \caption{Rate-distortion curves on the Kodak, the CLIC2020 and the MS-COCO 30K datasets.}
    \label{fig:rd_large}
\end{figure*}

We provide the additional KID \cite{binkowski2018demystifying} results on the CLIC2020 and the MS-COCO 30K datasets on the Fig. \ref{fig:exp_KID}. We present the evaluation results on the DIV2K dataset \cite{Agustsson_2017_CVPR_Workshops} at full resolution in Fig. \ref{fig:exp_DIV2k}. Large version of the rate-distortion curves is shown in Fig. \ref{fig:rd_large}. Diffusion-based methods \cite{careil2023towards, li2024extremeimagecompressionlatent} are excluded from evaluation on this dataset due to their large memory requirements, which exceed the capacity of our GPU (A100 with 40GB memory). From Fig. \ref{fig:exp_DIV2k}, we can see that our method outperforms MS-ILLM \cite{muckley2023improving} across all reference and no-reference perceptual metrics, demonstrating its effectiveness. Additionally, we evaluate traditional pixel-level distortion metrics, PSNR and MS-SSIM \cite{wang2003multiscale}, for a more comprehensive analysis, as shown in Fig. \ref{fig:exp_PSNR}. It is worth noting that at extreme low bitrates, pixel-level distortion becomes too severe (e.g., the PSNR of VTM drops below 25 dB), making these metrics less meaningful for evaluating visual quality. Although our model does not exhibit the best pixel-level distortion metrics, it still provides the most visually appealing reconstructions with high fidelity, as demonstrated in previous sections.

\subsection{Qualitative Results}
\begin{figure*}
    \centering
    \includegraphics[width=1.0\linewidth]{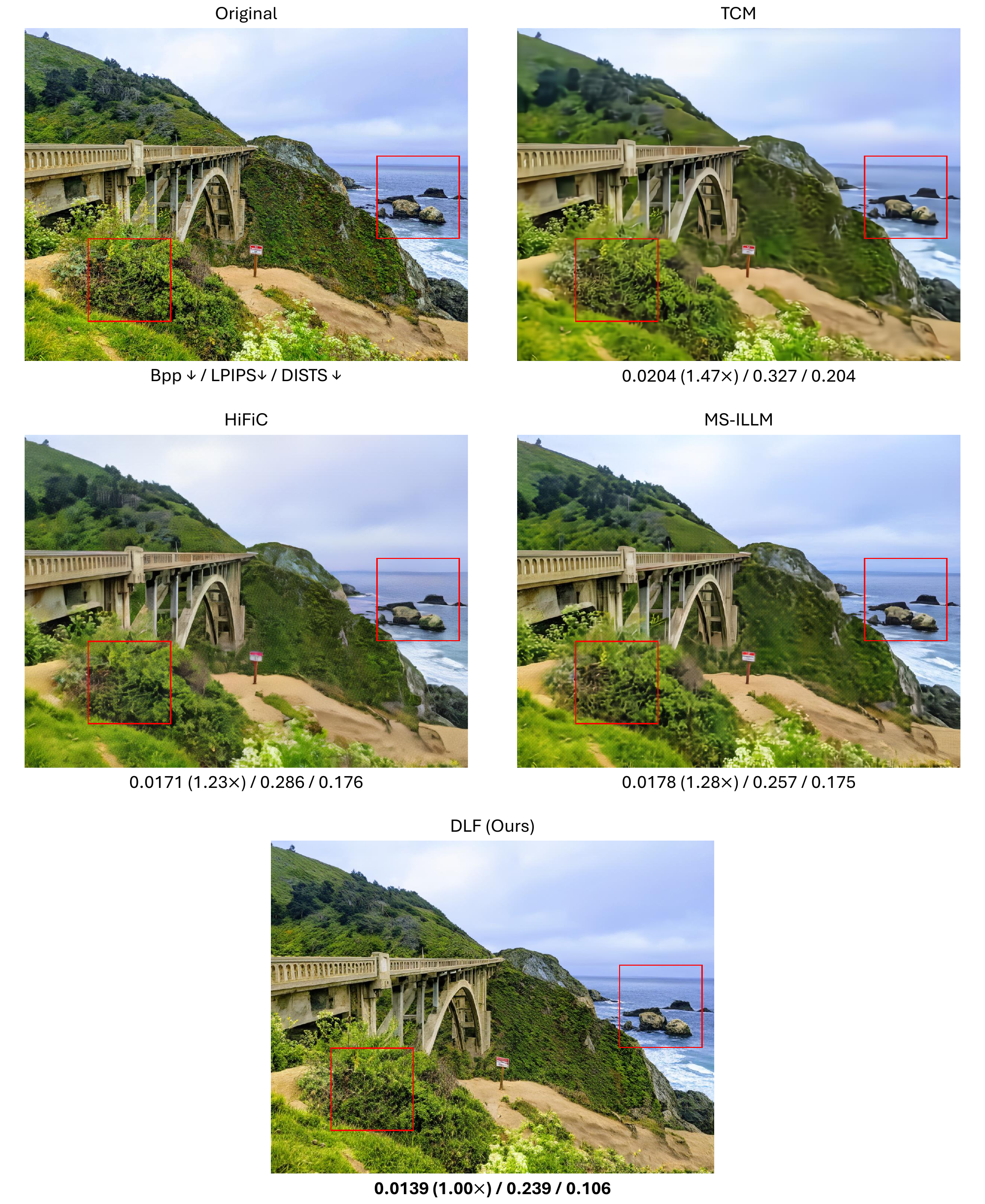}
    \vspace{-3mm}
    \caption{Qualitative examples on the CLIC2020 dataset (full resolution).}
    \label{fig:clic1}
\end{figure*}

\begin{figure*}
    \centering
    \includegraphics[width=1.0\linewidth]{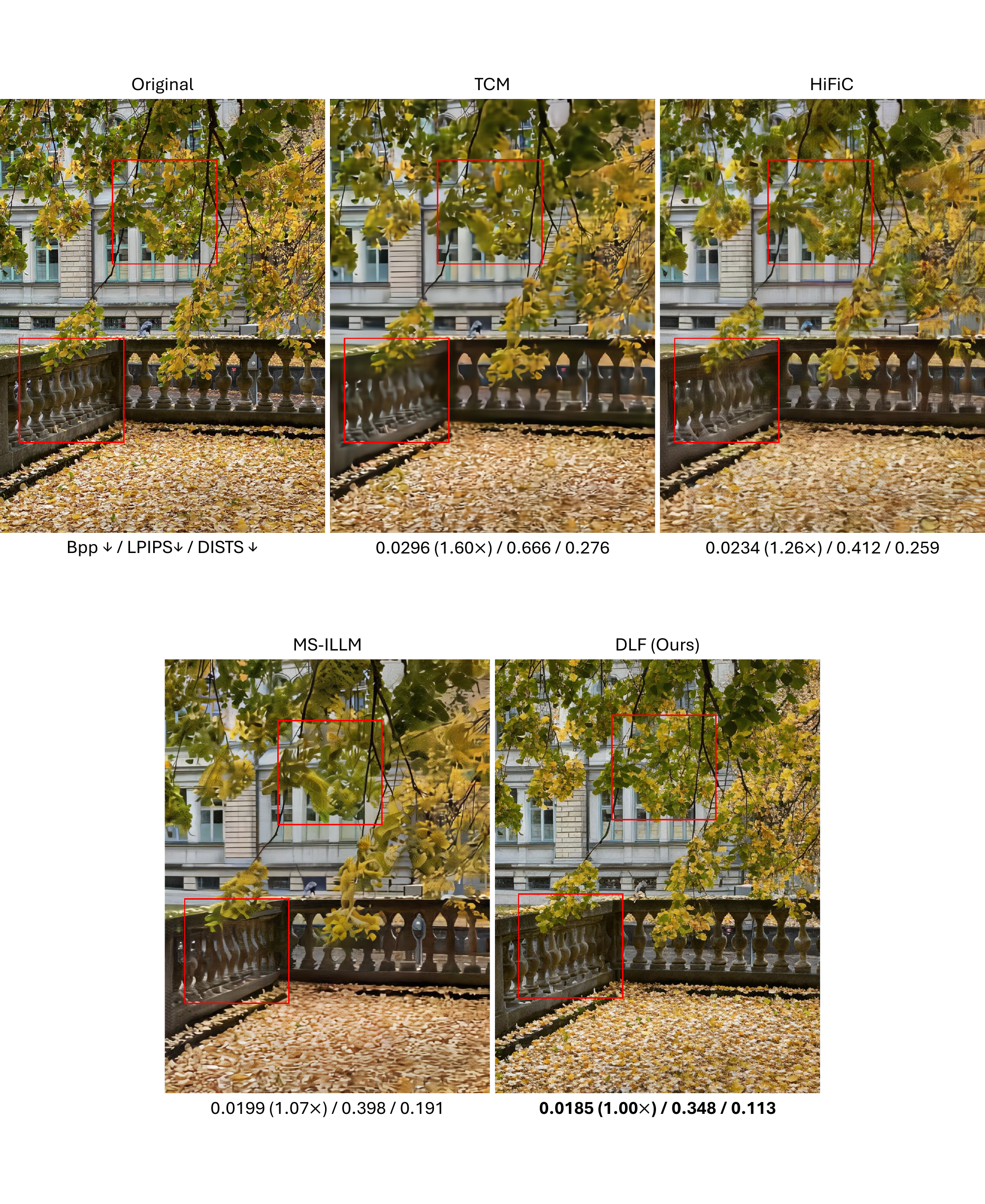}
    \vspace{-3mm}
    \caption{Qualitative examples on the CLIC2020 dataset (full resolution).}
    \label{fig:clic2}
\end{figure*}

\begin{figure*}
    \centering
    \includegraphics[width=1.0\linewidth]{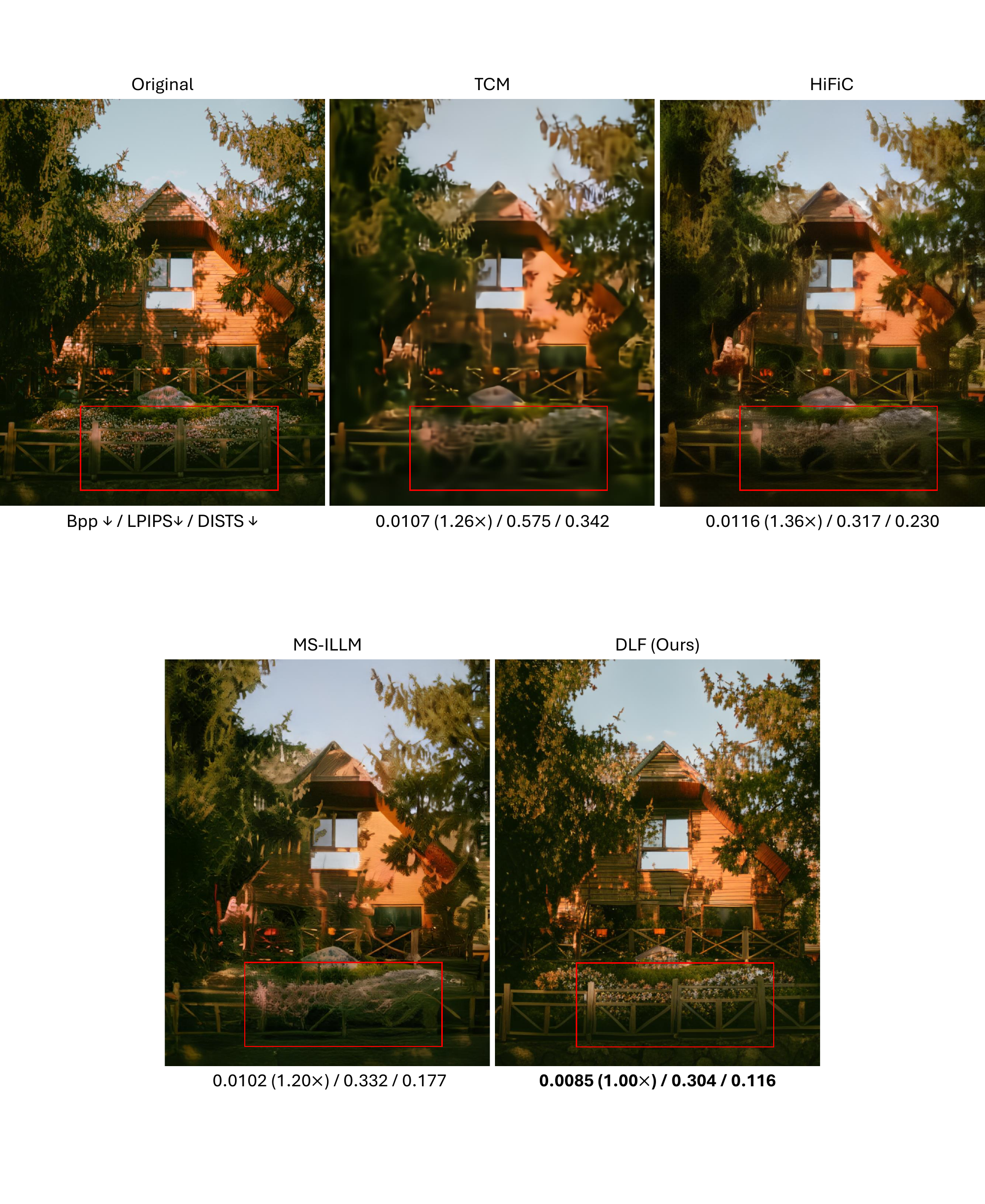}
    \vspace{-3mm}
    \caption{Qualitative examples on the CLIC2020 dataset (full resolution).}
    \label{fig:clic3}
\end{figure*}

\begin{figure*}
    \centering
    \includegraphics[width=1.0\linewidth]{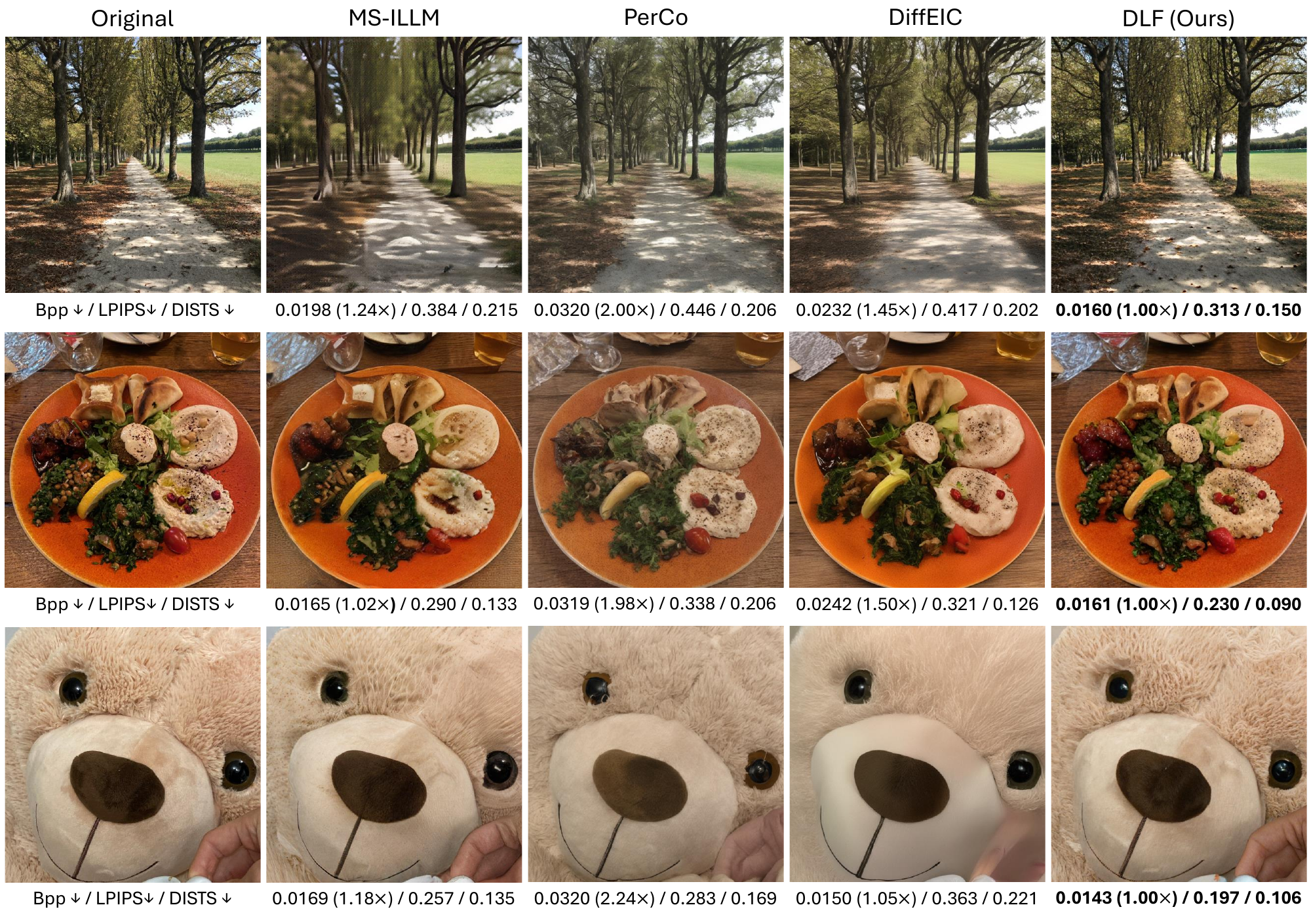}
    \vspace{-3mm}
    \caption{Qualitative examples on the CLIC2020 dataset (768$\times$768).}
    \label{fig:clic768}
\end{figure*}

\begin{figure*}
    \centering
    \includegraphics[width=1.0\linewidth]{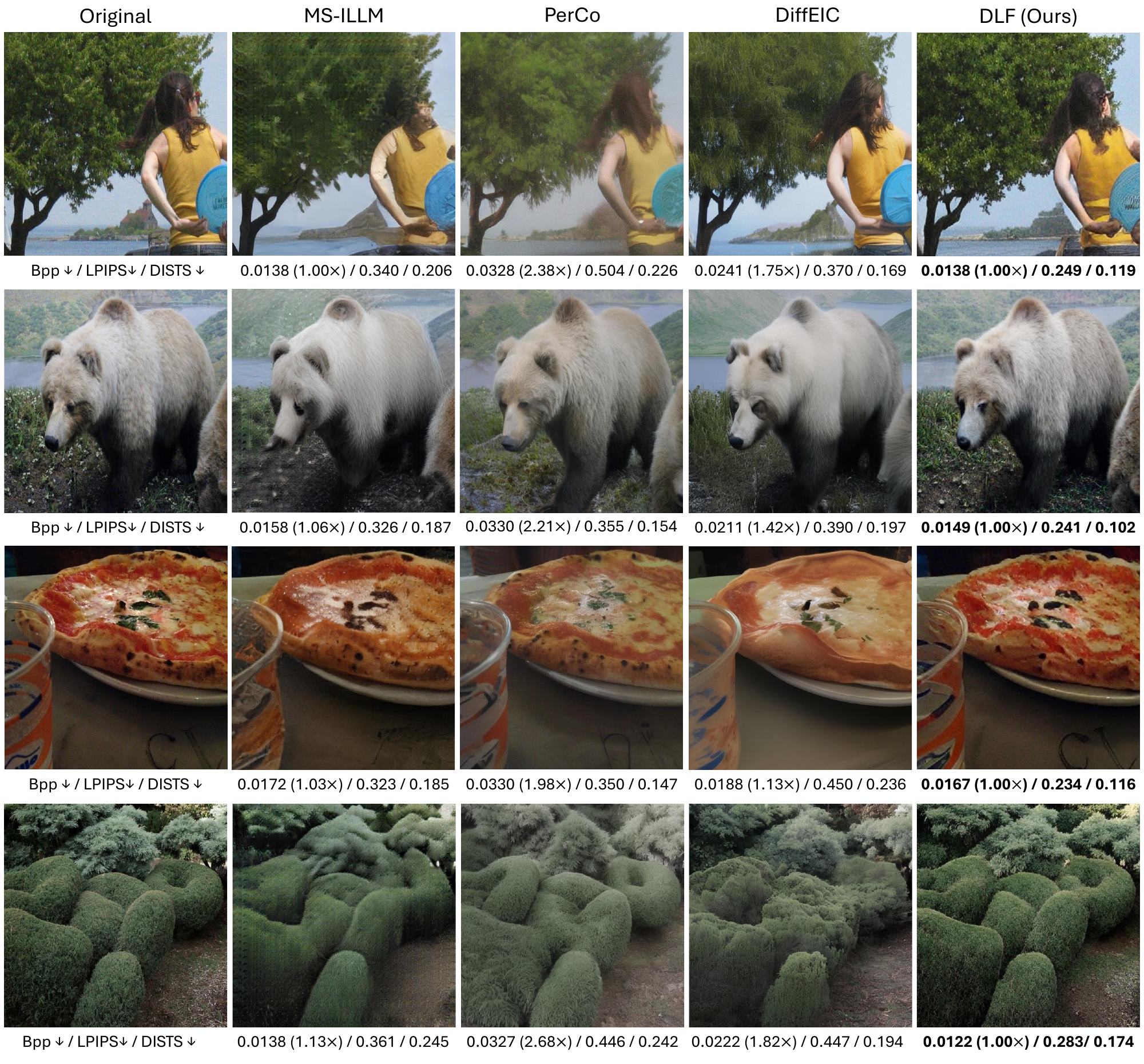}
    \vspace{-3mm}
    \caption{Qualitative examples on the MS-COCO 30K dataset.}
    \label{fig:coco}
\end{figure*}

In this section, we provide more visual examples across the CLIC2020 \cite{CLIC2020} (Full resolution: Fig. \ref{fig:clic1}, \ref{fig:clic2}, \ref{fig:clic3}; $768\times768$: Fig. \ref{fig:clic768}) and MS-COCO 30K \cite{lin2014microsoft} (Fig. \ref{fig:coco}) datasets. From these examples, we can find that DLF achieves the best quality with the lowest bitrate cost.

\end{document}